\documentclass[times,twocolumn,final,authoryear]{elsarticle}

\usepackage{ycviu}
\usepackage{framed,multirow}

\usepackage{amssymb}
\usepackage{latexsym}

\usepackage{url}
\usepackage{xcolor}
\definecolor{newcolor}{rgb}{.8,.349,.1}

\usepackage{booktabs} 
\usepackage{amsmath}

\journal{Computer Vision and Image Understanding}

\begin{document}
\thispagestyle{empty}                                             

\clearpage
\thispagestyle{empty}
\ifpreprint
  \vspace*{-1pc}
\fi

\clearpage
\thispagestyle{empty}

\ifpreprint
  \vspace*{-1pc}
\else
\fi

\clearpage

\ifpreprint
  \setcounter{page}{1}
\else
  \setcounter{page}{1}
\fi

\begin{frontmatter}

\title{ALCN: Adaptive Local Contrast Normalization}

\author[1]{Mahdi \snm{Rad}\corref{cor1}} 
\cortext[cor1]{Corresponding author: }
\ead{rad@icg.tugraz.at}
\author[1]{Peter M. \snm{Roth}}
\author[2,1]{Vincent \snm{Lepetit}}

\address[1]{Institute of Computer Graphics and Vision,
 Graz University of Technology, Graz, Austria}
\address[2]{LIGM, IMAGINE, Ecole des Ponts, Univ Gustave Eiffel, CNRS, Marne-la-Vall{\'e}e, France}


\begin{abstract}
To  make Robotics  and  Augmented Reality  applications  robust to  illumination
changes,  the current  trend is  to train  a Deep  Network with  training images
captured under many different lighting conditions.  Unfortunately, creating such
a training  set is a very unwieldy and complex task.   We therefore propose a  novel illumination
normalization  method  that can  easily  be  used  for different  problems  with
challenging  illumination conditions.   Our  preliminary  experiments show  that
among current normalization methods,  the Difference-of-Gaussians method remains
a very good baseline, and we  introduce a novel illumination normalization model
that generalizes it.  Our key insight  is then that the normalization parameters
should depend  on the input  image, and we aim  to train a  Convolutional Neural
Network to predict these parameters from the input image.  This, however, cannot
be done in a supervised manner, as  the optimal parameters are not known \emph{a
  priori}.  We thus designed a method to train this network jointly with another
network that aims to recognize objects under different illuminations: The latter
network  performs well  when the  former network  predicts good  values for  the
normalization  parameters. We  show  that our  method significantly  outperforms
standard normalization methods and would also be appear to be  universal since it does not have to
be re-trained for  each new application.  Our method improves  the robustness to
light changes  of state-of-the-art  3D object  detection and face  recognition methods.
\end{abstract}


\end{frontmatter}




\section{Introduction}

Over the last years, Deep Networks~(\cite{LeCun98,Krizhevsky12,Simonyan15}) have
spectacularly improved the performance of computer vision applications.  Development efforts to date, however,   have mainly been   focused  on tasks  where  large quantities  of
training  data are  available.   To  be robust  to  illumination conditions  for
example, one can  train a Deep Network with many  samples captured under various
illumination conditions.

\begin{figure*}
  \begin{center}
    \begin{tabular}{cc@{ }c@{ }c}
      \includegraphics[width=0.225\linewidth]{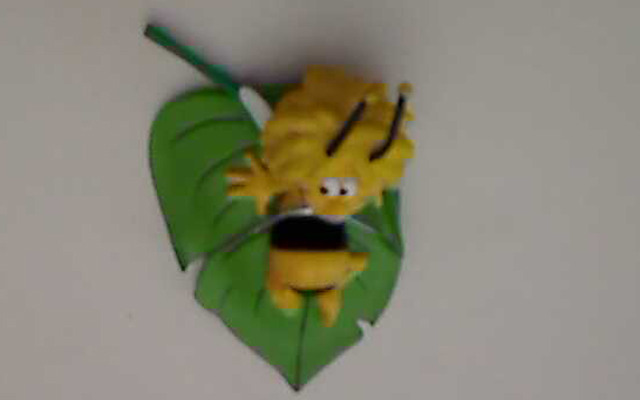} &
      \includegraphics[width=0.225\linewidth]{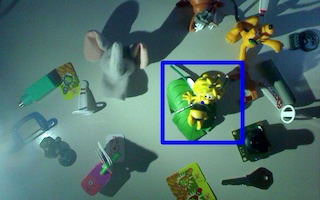} &
      \includegraphics[width=0.225\linewidth]{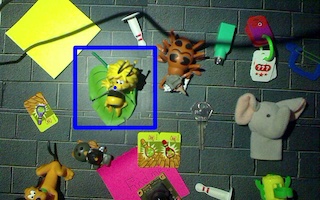} &
      \includegraphics[width=0.225\linewidth]{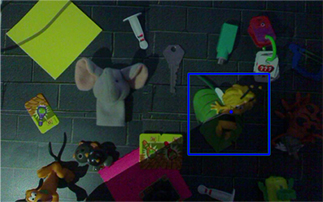}
      \\
      \includegraphics[width=0.225\linewidth]{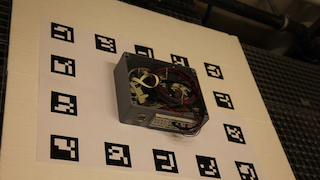} &
      \includegraphics[width=0.225\linewidth]{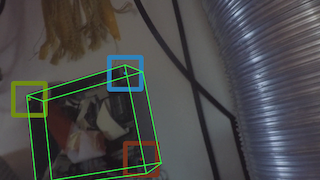} &
      \includegraphics[width=0.225\linewidth]{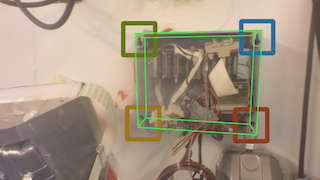} &
      \includegraphics[width=0.225\linewidth]{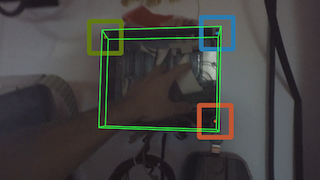}
      \\
      \includegraphics[width=0.225\linewidth]{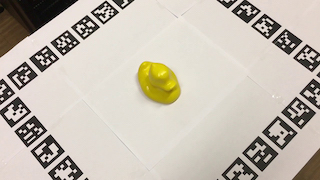} &
      \includegraphics[width=0.225\linewidth]{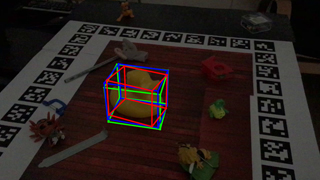} &
      \includegraphics[width=0.225\linewidth]{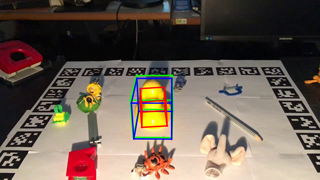} &
      \includegraphics[width=0.225\linewidth]{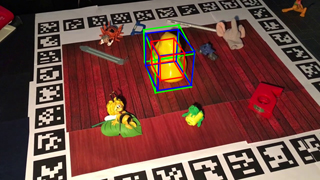}
      \\
      \multicolumn{1}{c}{Training Images} & \multicolumn{3}{c}{Test Images}\\
    \end{tabular}
  \end{center}
  \caption{\label{fig:teaser}  We  propose  a  novel  approach  to  illumination
    normalization, which allows  us to deal with strong light  changes even when
    only few training samples are available.  We apply it to 2D detection (first row)
    and  3D object  detection using  the methods  of~\cite{Crivellaro15} (second
    row) and  of \cite{bb8}  (third row): Given  training images  under constant
    illumination, we  can detect the object  and predict its pose  under various
    and drastic illumination.   In the third row, green bounding  boxes show the
    ground  truth pose,  blue bounding  boxes represent  the pose  obtained with
    ALCN, and red bounding boxes  the pose obtained without normalization. }
\end{figure*}

While for some general categories such  as faces, cars, or pedestrians, training
data can be exploited from other  data, or the capturing of many images under different
conditions is also possible, these processes become very unwieldy and complex tasks for others.  For example, as
illustrated  in  Fig.~\ref{fig:teaser}, we  want  to  estimate  the 3D  pose  of
specific objects without having to vary the illumination when capturing training
images.  To achieve  this, we could use a contrast  normalization technique such
as Local  Contrast Normalization~(\cite{Jarrett09}),  Difference-of-Gaussians or
histogram normalization.   Our  experiments show, however, that  existing methods
often fail when dealing with large magnitudes of illumination changes.

Among  the  various  existing normalization  methods,  Difference-of-Gaussians
still performs  best in our  experiments, which inspired us to introduce a 
normalization model building on a  linear combination  of 2D  Gaussian 
kernels  with fixed standard deviations.  But instead of using fixed parameters, 
we propose to adapt these parameters to the illumination  conditions of the 
different image regions: By this means, we can handle bigger illumination changes and 
avoid manual tuning.

However,  the  link  between a  given  image  and  the  best parameters  is  not
straightforward.  We  therefore want to  learn to predict these  parameters from
the image using a  CNN.  Since we do not have \emph{a  priori} knowledge-the parameters to
predict, we cannot train this CNN in a standard supervised manner.  Our solution
is to train it  \emph{jointly} in a supervised way together  with another CNN to
achieve object detection under illumination changes.

We  call this  method Adaptive  Local  Contrast Normalization~(ALCN),  as it  is
related to previous  Local Contrast Normalization methods  while being adaptive.
We  show  that  ALCN  outperforms    previous  methods  for
illumination  normalization by   a  large  margin  while  we  do  not need  any manual  tuning.   It  
also
outperforms     Deep     Networks    including     VGG~(\cite{Simonyan15})     and
ResNet~(\cite{he2016deep}) trained on  the same images, showing  that our approach
can generalize better with unseen illumination variations than a single network.

In summary, our  main contribution  is an  efficient method  that makes  Deep
Networks more robust  to illumination changes that have not been seen during training, therefore
requiring  much far less  training  data.  Furthermore,  we created  new
datasets  for benchmarking  of object  detection  and 3D  pose estimation  under
challenging  lightening   conditions  with  distractor  objects   and  cluttered
background.

We published a first version of this work in~\cite{rad2017alcn}. This paper
extends this work in the following manner:

\begin{itemize}
  \item We provide an extensive overview of existing normalization methods.
  \item We perform thorough ablation studies to justify our contribution.
  \item We also perform experiments on network design and the impact of different activation functions.
  \item We evaluate our normalization method on other applications such as 3D object detection and pose estimation and  face recognition, which our approach was not trained for.
\end{itemize}

In the remainder of this paper, we first discuss related work in Section~\ref{sec:related_work}, we then review the existing normalization methods and introduce our normalization model in Section~\ref{sec:adaptive}, and we evaluate it on different applications in Section~\ref{sec:experiments}.

\section{Related Work}
\label{sec:related_work}

Reliable computer vision methods need to be invariant, or at least robust, to
many different visual nuisances, including  pose and illumination variations. In 
the following, we give an overview of the different,  and sometimes complementary 
approaches for achieving  this.

\paragraph{Image normalization methods} A  first approach  is  to  normalize the  input  image  using image  statistics.
Several methods have  been proposed, sometimes used together  with Deep Networks
such as SLCN and DLCN: Difference-of-Gaussians~(DoG), Whitening, Subtractive and
Divisive Local Contrast  Normalization~(SLCN and DLCN)~(\cite{Jarrett09}), Local
Response Normalization~(LRN)~(\cite{Krizhevsky12}), Histogram Equalization~(HE),
Contrast Limited  Adaptive Histogram  Equalization~(CLAHE)~(\cite{Pizer87}).  We
detail these methods  in Section~\ref{sec:overview_existing_normalizations}, and
compare to them in our experiments  in  Section~\ref{sec:results}.

However, illumination is not necessarily uniform  over an image: Applying one of
these methods  locally over  regions of  the image  handles local  light changes
better,  but unfortunately  they can  also  become unstable  on poorly  textured
regions.  Our  approach overcomes this  limitation with an adaptive  method that
effectively adjusts the  normalization according to the local  appearance of the
image.

\paragraph{Invariant features} An alternative  method is to use  locally invariant features. For  example, Haar
wavelets~(\cite{Viola04})  and the  pairwise intensity  comparisons used  in Local
Binary  Patterns~(\cite{Ojala02})  are  invariant  to  monotonic  changes  of  the
intensities.  Features based on image gradients are invariant to constants added
to the  intensities. In practice, they  are also often made  invariant to affine
changes  by normalizing  gradient  magnitudes  over the  bins  indexed by  their
orientations~(\cite{Levi04}).   The  SIFT   descriptors  are  additionally
normalized by an iterative process that  makes them robust to saturation effects
as well~(\cite{Lowe04}).  However,  it is difficult to come up  with features that
are invariant to complex illumination changes  on 3D objects, such as changes of
light direction,  cast or self  shadows.

\paragraph{Intrinsic images} A third approach is to model  illumination explicitly and estimate an intrinsic
image  or  a  self quotient  image  of  the  input  image,  to get  rid  of  the
illumination  and isolate  the  reflectance  of the  scene  as  an invariant  to
illumination~(\cite{Wang04,Shen11,Shen11b,shen2013intrinsic,Zhou15,nestmeyer2017reflectance,Fan_2018_CVPR,bi20151}).    However,   it  is   still
difficult to  get an intrinsic  image from one single  input image that  is good
enough    for    computer    vision     tasks,    as    our    experiments    in
Section~\ref{sec:results} will show for 2D object detection.

\paragraph{Data-driven robustness} The current trend to achieve robustness  to illumination changes is to train Deep
Networks    with   different    illuminations    present    in   the    training
set~(\cite{Simonyan15,he2016deep,huang2017densely,xie2017aggregated}).   This,  however,  requires  the  acquisition  of  many
training  images  under various  conditions.   As  we  will show,  our  approach
performs  better than  single  Deep Networks  when  illumination variations  are
limited  in the  training  set, which  can  be  the case  in  practice for  some
applications.


\newcommand{\imageref}{I_\text{ref}}
\newcommand{\imagenew}{I_\text{new}}
\newcommand{\imageint}{I_\text{interm}}

\newcommand{\bg}{\text{bg}}
\newcommand{\clip}{\text{clip}}
\newcommand{\scale}{\text{scale}}

\newcommand{\GN}{\text{NS}}
\newcommand{\DoG}{\text{DoG}}
\newcommand{\LCN}{\text{LCN}}
\newcommand{\ALCN}{\text{ALCN}}
\newcommand{\DLCN}{\text{DLCN}}
\newcommand{\SLCN}{\text{SLCN}}
\newcommand{\DIV}{\text{Div}}
\newcommand{\SUB}{\text{Sub}}
\newcommand{\Ret}{\text{DoG}}
\newcommand{\LRN}{\text{LRN}}

\section{Adaptive Local Contrast Normalization}
\label{sec:adaptive}

In  this section,  we  first  provide an  overview  of the  existing
normalization methods,  since we  will compare our  method against  them in
  Section~\ref{sec:alcn_vs_normalization_methods}.    We  then   introduce  our
normalization model,  then we discuss  how we train a  CNN to predict  the model
parameters  for a  given image  region and  how we  can efficiently  extend this
method to a whole image.

\subsection{Overview of Existing Normalization Methods}
\label{sec:overview_existing_normalizations}

We describe below  the main different existing methods to  make object detection
techniques  invariant to  light  changes. These  are also  the  methods we  will
compare to  in Section~\ref{sec:results}.  \emph{In  practice, in our  2D object
  experiments presented  below, we use a  detector in a sliding  window fashion,
  and we apply these normalization methods,  including our Adaptive LCN, to each
  window independently. }

\begin{itemize}
\item \textbf{Normalization by Standardization~(NS)}. A common method to be robust to
  light changes is to replace the input image $I$ by:
  \begin{equation}
    I^\GN = \frac{(I - \bar{I})}{\sigma_I} \> ,
  \end{equation}
  where  $\bar{I}$  and  $\sigma_I$  are  respectively  the  mean  and  standard
  deviation  of the  pixel intensities  in $I$.   This transformation  makes the
  resulting  image window  $I^\GN$  invariant to  affine  transformation of  the
  intensities---if we ignore  saturation effects that clip  the intensity values
  within $[0;255]$.

\item  \textbf{Difference-of-Gaussians~(DoG)}.  The Difference-of-Gaussians  is  a
  band-pass filter often used for normalization:
  \begin{equation}
    I^\DoG = (k_2^\DoG\cdot G_{\sigma^\DoG_2} - k_1^\DoG \cdot G_{\sigma^\DoG_1})*I \> ,
    \label{eq:dog}
  \end{equation}
  where $G_\sigma$ is  a 2D Gaussian filter of standard  deviation $\sigma$, and
  $k_1$,  $k_2$,  $\sigma_1$,   $\sigma_2$  are  parameters.   $*$   is  the  2D
  convolution operator.   This is also a  common mathematical model for  the ON-
  and  OFF-center  cells  of  the retina~(\cite{Dayan05}).  In  practice,  we  use
  Gaussian filters of size $\lceil 6 \sigma + 1 \rceil$, to truncate
  only very small values of the Gaussian kernels.

  \newcommand{\bC}{{\bf C}}
  \newcommand{\bD}{{\bf D}}
  \newcommand{\bE}{{\bf E}}
  \newcommand{\bW}{{\bf W}}
  
\item  \textbf{Whitening}.   Whitening  is  sometimes  used  for  illumination
  normalization.  It is  related to DoG as learned  whitening filters computed
  from         natural         image        patches         resemble         a
  Difference-of-Gaussians~(\cite{Rigamonti11b,maxout,Yang15,Paulin15}).     In
  practice, we first compute the whitening matrix as the inverse of the square
  root of  the covariance  matrix of  the image patches.   The columns  of the
  whitening matrix are  all translated versions of the same  patch, and we use
  the      middle     column      as      the     whitening      convolutional
  filter~(\cite{Rigamonti11b}).

  \newcommand{\bm}{{\bf m}}
  
\item \textbf{Local  Contrast Normalization~(LCN)}.  When working with  Deep Networks,
  Local Contrast  Normalization~(LCN)~(\cite{Jarrett09}) is often used.   We tried
  its two  variants. Subtractive LCN  is also closely  related to DoG  as it
  subtracts from  every value in an  image patch a Gaussian-weighted  average of
  its neighbors:
  \begin{equation}
    I^{\SLCN} = I - G_{\sigma^\SUB} * I \> ,
    \label{eq:sub}
  \end{equation}
  where $\sigma^\SUB$ is  a parameter.  Divisive LCN, the  second variant, makes
  the image  invariant to local  affine changes  by dividing the  intensities in
  $I^{\SLCN}$ by their standard deviation, computed locally:
  \begin{equation}
    I^{\DLCN}(\bm) = \frac{I^{\SLCN}(\bm)}
    {\max\!\Big(t, \big(G_{\sigma^\DIV}*(I^{\SLCN})^2\big)(\bm)\Big)} , 
    \label{eq:div}
  \end{equation}
  where  $(I^{\SLCN})^2$ is  an  image  made of  the  squared intensities  of
  $I^{\SLCN}$, and $\sigma^\DIV$  is a parameter controlling  the size of
  the region for the local standard deviation of the intensities. $t$ is a small
  value to avoid singularities.
  
\item    \textbf{Local   Response    Normalization~(LRN)}.   Local    Response
  Normalization is  related to LCN,  and is also  used in many  applications to
  normalize     the    input     image,     or    the     output    of     the
  neurons~(\cite{Krizhevsky12,badrinarayanan2015segnet}).  The normalized value
  at location $\bm$ after applying kernel $i$ can be written as:
  \begin{equation}
    \label{eq:lrn}
    I^{\LRN}_{(i)}(\bm) = \frac{I_{(i)}(\bm)}{\left( k + \alpha \sum_{j=\max(0, i-n/2)}^{\min(N-1, i+n/2)} (I_{(j)}(\bm))^2\right)^\beta}
  \end{equation}
  where the sum is  over the $n$ kernel maps around index $i$,  and $N$ is the
  total number of  kernels in the layer. Constants $k,  n, \alpha$ and $\beta$
  are then manually  selected.  Compared to LCN, LRN aims  more at normalizing
  the image in terms of brightness rather than contrast~(\cite{Krizhevsky12}).

\item \textbf{Histogram Equalization~(HE)}.  Histogram Equalization aims at enhancing
  the image contrast by better distributing  the intensities of the input image.
  First, a histogram  $p(\lambda_i)$ of the image  intensities, with $\lambda_i$
  any  possible quantized  intensity value,  is  built.  Then,  a new  intensity
  $\tilde\lambda_i$ is  assigned to all  the pixels with  intensity $\lambda_i$,
  with
  \begin{equation}
    \label{eq:histEq}
    \tilde\lambda_i = \lambda_{\min} + \textrm{floor}\Big((\lambda_{\max}-\lambda_{\min})\sum_{j=0}^i
    p(\lambda_j) \Big) \> .
  \end{equation}
  
\item \textbf{Contrast Limited  Adaptive Histogram Equalization~(CLAHE)}.  While
  Histogram Equalization does  not take the spatial location of  the pixels into
  account, CLAHE~(\cite{Pizer87})  introduces spatial constraints and  attempts to
  avoid noise amplification: It performs Histogram equalization locally, and the
  histograms  are  clipped:  If  $p(\lambda_i)$   is  higher  than  a  threshold
  $\hat\lambda$, it is set to $\hat\lambda$ and the histogram is re-normalized.

\item \textbf{Intrinsic Image}.  An intrinsic image of an input  image $I$ can
  be obtained by  separating the illumination $S$ from the  reflectance $R$ of
  the scene:
  \begin{equation}
    \label{eq:intrinsicImage}
    I(\bm) = S(\bm) R(\bm) \> .
  \end{equation}
  Eq.~\eqref{eq:intrinsicImage} is ill-posed, but can be solved by adding
  various constraints~(\cite{Shen11,Shen11b,Zhou15}). Since $R$ is supposed to be
  free from illumination effects,  it can then be used as  input instead of the
  original  image to  be  invariant  to illuminations.   However,  it is  still
  difficult to estimate $R$ robustly, as our experiments will show. Moreover,
  optimizing over $S$ and $R$ under constraints is computationally expensive,
  especially for real-time applications.
  
\item    \textbf{Self     Quotient    Image~(SQI)}.     The     Self    Quotient
  Image~(\cite{Wang04}) aims at estimating the object reflectance field from a 2D
  image similarly to the Intrinsic Image  method, but is based on the Lambertian
  model instead of the reflectance illumination model. The Self  Quotient Image $Q$
  of image $I$ is defined by:
  \begin{equation}
    \label{eq:sqi}
    Q = \frac{I}{G_\sigma^\text{SQI} * I} \> .
  \end{equation}
  
\end{itemize}

\subsection{Normalization Model}
\label{sec:normalization_model}
As   our   experiments   in   Section~\ref{sec:experiments}   will   show,   the
Difference-of-Gaussians normalization  method performs  best among  the existing
normalization methods, however, it is difficult to find the standard deviations
that perform well for any input image, as we will discuss in Section~\ref{sec:different_images_need_different_parameters}. We therefore introduce the following
formulation for our ALCN method:
\begin{equation}
  \ALCN(I; \;w) = \left(\sum_{i=1}^N w_i\;\cdot \;G_{\sigma_i^\ALCN}\right)*I \> ,
  \label{eq:AdaptiveLCN}
\end{equation}
where $I$ is  an input image window,  $w$ a vector containing  the parameters of
the method  and $\ALCN(I;  w)$ is  the normalized  image; $G_\sigma$  denotes a
Gaussian kernel of  standard deviation $\sigma$; the  $\sigma_i^\ALCN$ are fixed
standard  deviations,  and   $*$  denotes  the  convolution   product.   In  the
experiments, we  use ten  different 2D Gaussian  filters $G_{\sigma_{i}^\ALCN}$,
with standard deviation $\sigma_{i}^\ALCN = i /  2$ for $i=1, 2, ..., 10$.  This
model  is  a generalization  of  the  Difference-of-Gaussians model,  since  the
normalized image is obtained by convolution  of a linear combination of Gaussian
kernels, and  the weights of  this linear combination  are the parameters  of the
model.

Using fixed 2D Gaussian filters allows us to have fast
  running time: During training, we can perform
  the Gaussian convolutions on the samples of the mini-batches. It
  also makes training easier since the network has to predict only the
  weights of a linear combination. During testing, this allows us to
  efficiently varies the model parameters with the image locations
  efficiently, as will be explained in Section~\ref{sec:imageNormalization}.

\subsection{Joint Training to Predict the Model Parameters}
\label{sec:jointTraining}

\begin{figure*}[t]
  \begin{center}
    \begin{tabular}{cc}
      \includegraphics[trim=0cm 5.8cm 0cm 4.0cm,clip=true,width=0.48\linewidth]{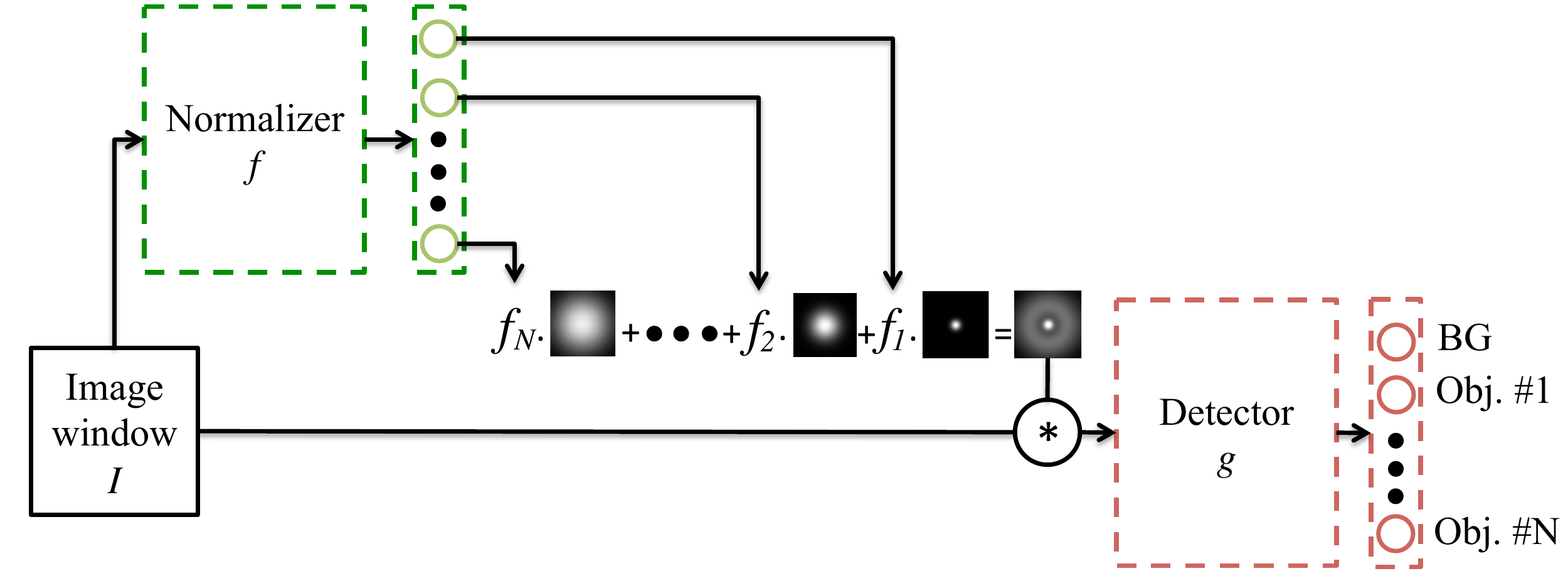} &
      \includegraphics[trim=2cm 3.8cm 1.5cm 4.0cm,clip=true,width=0.48\linewidth]{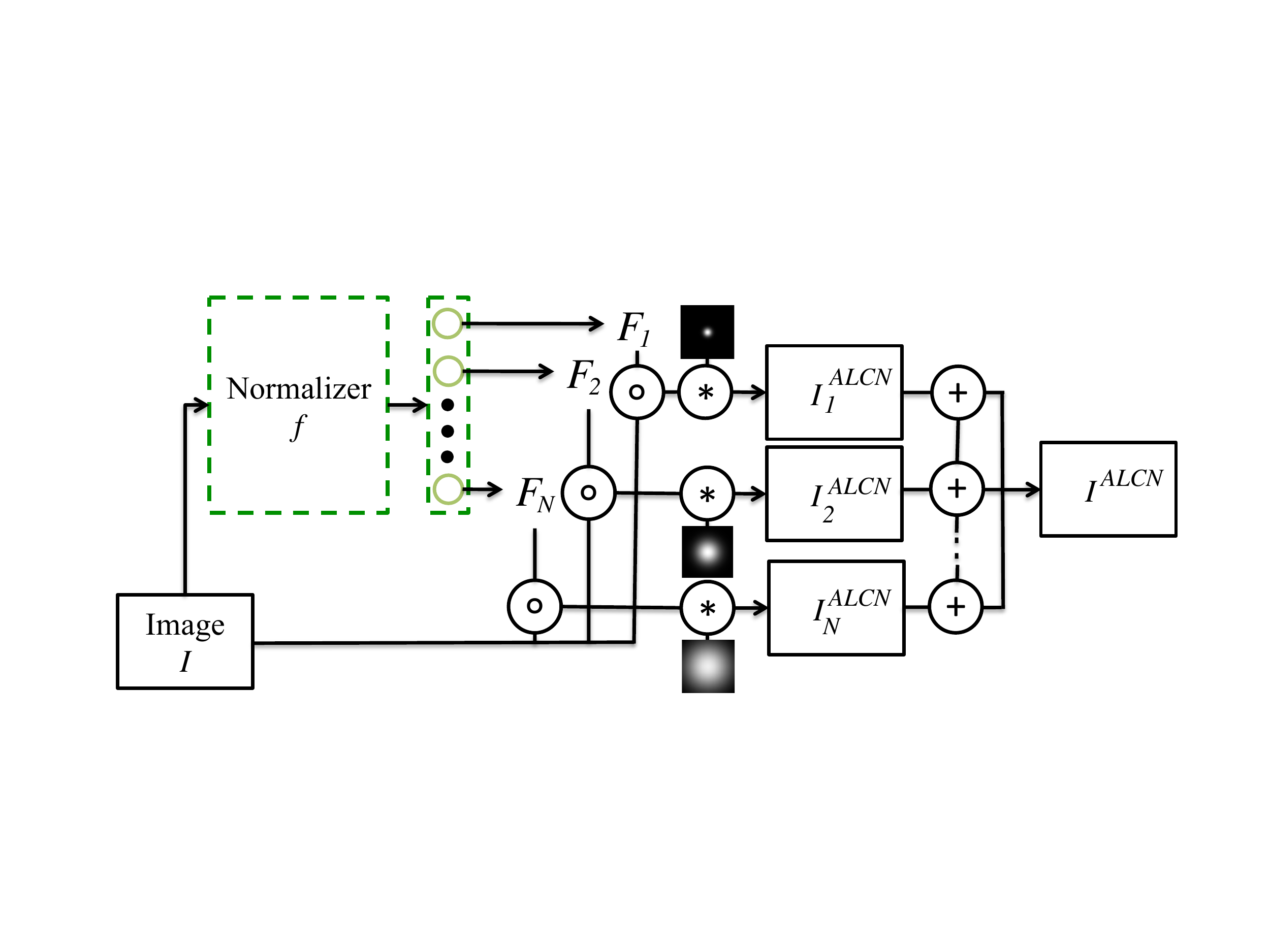} \\
      (a) & (b)\\
    \end{tabular}
  \end{center}
  \caption{\label{fig:system_overview}  Overview of  our method.   (a) We  first
    train our Normalizer jointly with the Detector using image regions from the
    Phos dataset. (b) We can then  normalize images of previously unseen objects
    by applying  this Normalizer to predict the parameters of our normalization
    model.
  }
\end{figure*}

As discussed in the introduction and shown in Fig.~\ref{fig:system_overview}(a),
we train a  Convolutional Neural Network~(CNN) to predict the  parameters $w$ of
our model for a  given image window $I$, jointly with  an object classifier.  We
call this CNN the Normalizer.

\begin{figure*}[t]
  \begin{center}
    \begin{tabular}{ccccc}
      \includegraphics[width=0.16\linewidth]{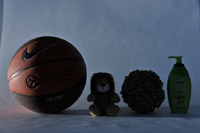}  &
      \includegraphics[width=0.16\linewidth]{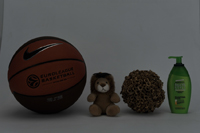} &
      \includegraphics[width=0.16\linewidth]{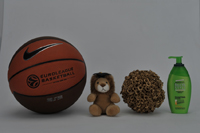} &
      \includegraphics[width=0.16\linewidth]{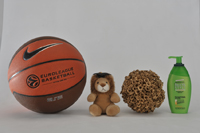} &
      \includegraphics[width=0.16\linewidth]{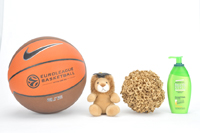} \\ 
    \end{tabular}
  \end{center}
  \caption{Four of the ten objects  we use from the Phos dataset~(\cite{phos2013}) under  different  illuminations.}
  \label{fig:phosDataset}

\end{figure*}

Like the  Normalizer, the classifier  is also implemented as a  CNN as well,  since deep
architectures  perform  well for  such  problems.   This  will also  make  joint
training of the Normalizer and the classifier easy.  We refer to this classifier
as  the Detector.   Joint  training is  done by  minimizing  the following  loss
function:
\begin{equation}
  (\hat{\Theta}, \hat{\Phi}) =
  \arg\min_{\Theta, \Phi} \sum_j
  \ell\left(g^{(\Theta)}\left(\ALCN(I_j;\;f^{(\Phi)}(I_j))\right);\; y_j\right) \> ,
  \label{eq:joint}
\end{equation}
where $\Theta$ and $\Phi$ are the parameters  of the Detector CNN $g(\cdot)$ and the
Normalizer $f(\cdot)$, respectively; $\ell(\cdot; y)$ is the negative log-likelihood loss
function.  $I_j$  and $y_j$ are training image regions  and their labels: We
use image regions extracted from the Phos dataset~(\cite{phos2013}), 
including the
images shown in Fig.~\ref{fig:phosDataset}, the  labels are either background or
the index of the object contained in the corresponding image region.  We use 
Phos  for our  purpose because  it is  made of  various objects  under different
illumination  conditions, with  9  images captured  under  various strengths  of
uniform illumination and  6 images under non-uniform  illumination from various
directions.   In   practice,  we  use   Theano~(\cite{bergstra_al:2010-scipy})  to
optimize Eq.~\eqref{eq:joint}.

\subsection{Different Images Need Different Parameters}
\label{sec:different_images_need_different_parameters}
In order to  show that  different images  need different  parameters when  using previous
normalization methods,  we performed two  studies. For each study we jointly optimizing the
four DoG parameters, at the same time as the Detector:
\begin{equation}
  (\hat{\Theta}, \hat{\Omega}) =
  \arg\min_{\Theta, \Omega} \sum_i
  \ell\left(g^{(\Theta)}\left(\DoG^{(\Omega)}*I_i\right); y_i\right) \> ,
  \label{eq:param_optimization}
\end{equation}
where $\Theta$  and $\Omega$ are the  parameters of the Detector  CNN $g(\cdot)$ and
\DoG~respectively, the  $\{(I_i,y_i)\}_i$ are annotated training  image windows,
and $\ell(\cdot;  y)$ is  the negative  log-likelihood loss  function. 

\paragraph{\bf{Effect of Brightness}}

In order to evaluate the effect of brightness, we split  the training
set into dark and bright images,  by simply thresholding the mean intensity, and
training two different  detectors, one for each subset,
We will give more details of the dataset in Section~\ref{sec:datasets}.

We set the intensity threshold to 80  in our experiments.  At run-time, for each
possible image  location, we first  tested if the  image patch centered  on this
location is  dark or bright, and  apply the corresponding CNN.   We also trained
two  more detectors,  one for  each  subset, but  this time  with the  optimized
parameter values obtained on the other subset.

\begin{figure*}
  \begin{center}
    \begin{tabular}{ccc}
      \includegraphics[trim=0.7cm 0.1cm 1.5cm 0.05cm,clip=true, width=0.30\linewidth]{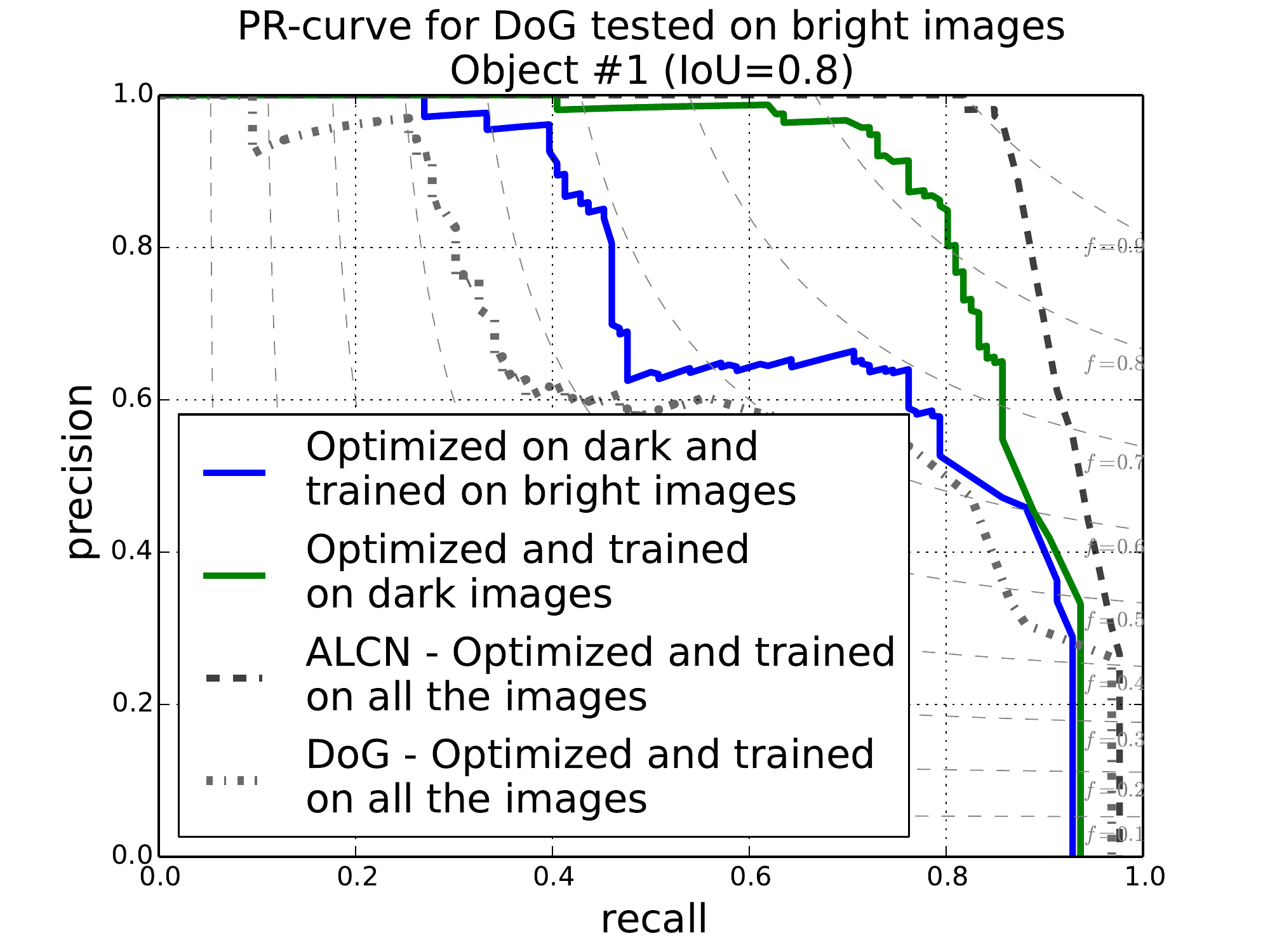} &
      \includegraphics[trim=0.7cm 0.1cm 1.5cm 0.05cm,clip=true, width=0.30\linewidth]{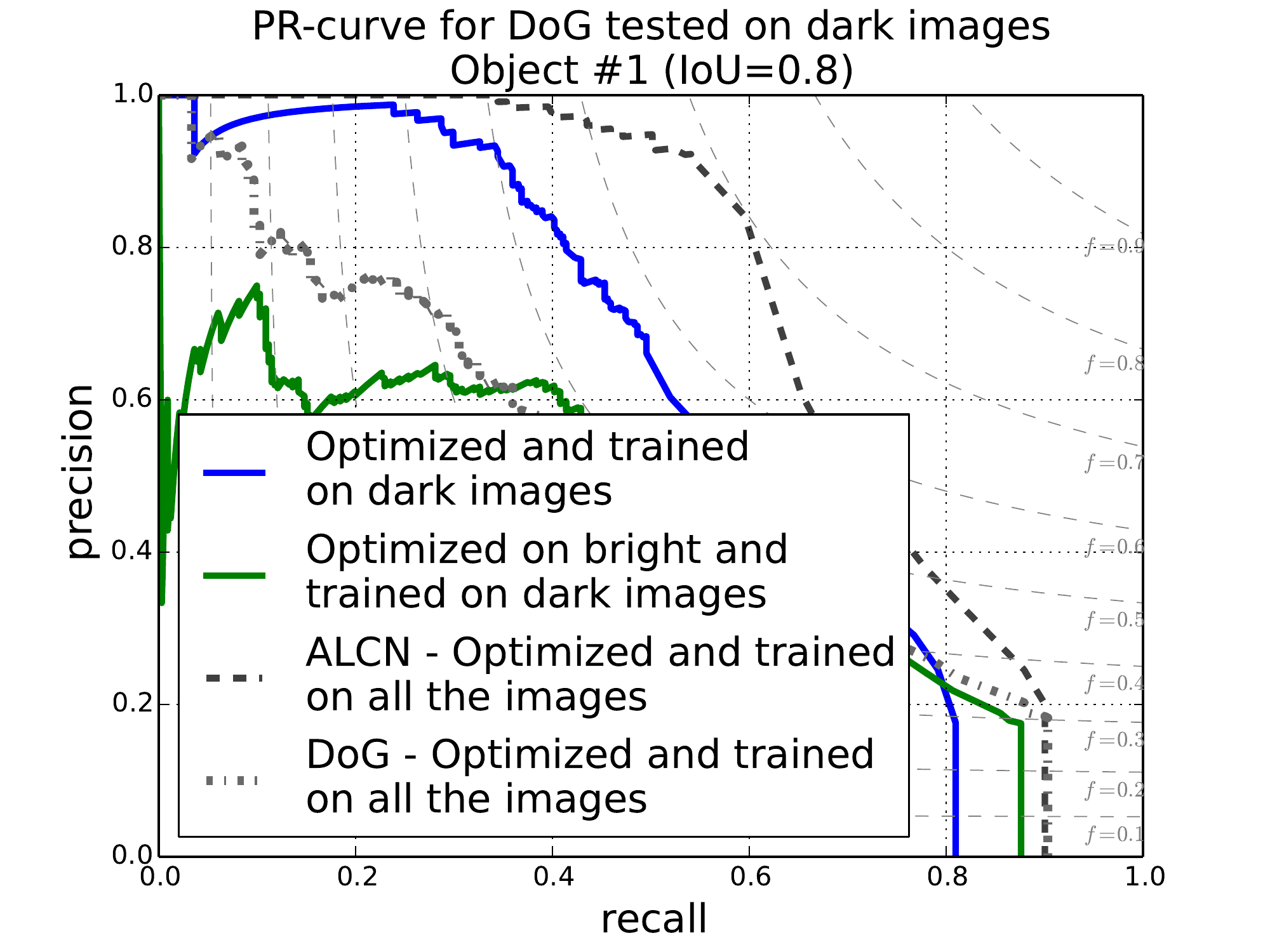} &
      \includegraphics[trim=0.7cm 0.1cm 1.5cm 0.05cm,clip=true, width=0.30\linewidth]{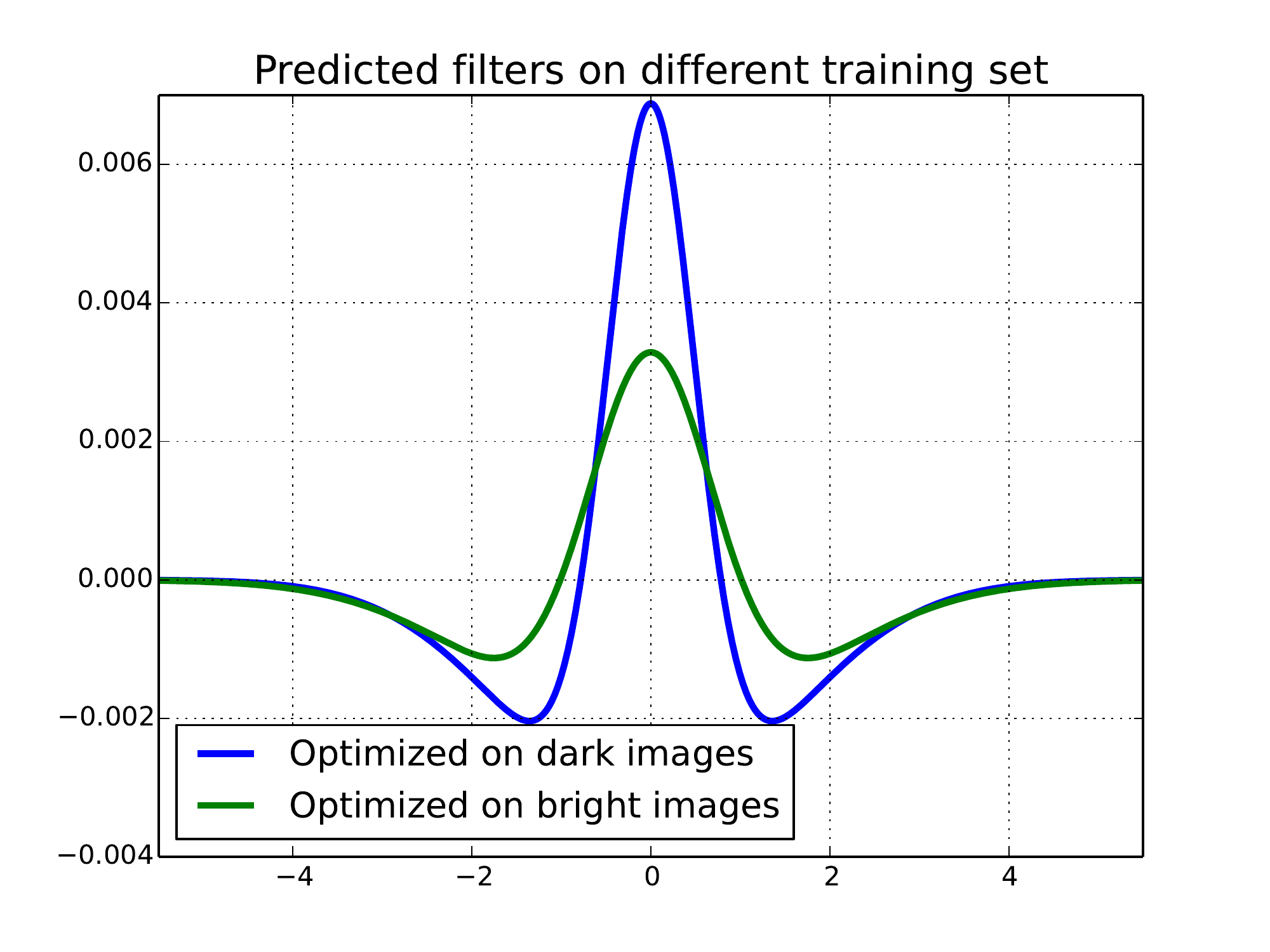}\\
    \end{tabular}
  \end{center}
  \caption{Evaluating   the  relations   between   brightness  and   optimized
    normalization parameter values.  We split our dataset into bright and dark
    images. The  best performance on the  dark images is obtained  with larger
    filters than for the bright images.}
  \label{fig:Retina_parameterization}
\end{figure*}

\begin{figure*}
 \begin{center}
   \begin{tabular}{cc}
     \begin{tabular}{cccc}
       \includegraphics[width=0.16\linewidth]{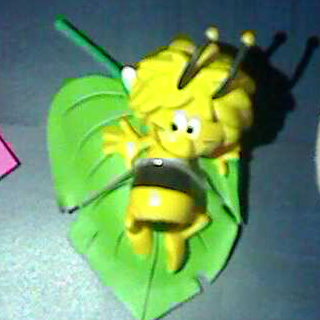} & \hspace{-3mm}
       \includegraphics[width=0.16\linewidth]{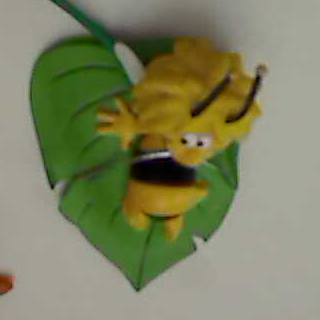} &
       \hspace{-3mm}
       \includegraphics[width=0.16\linewidth]{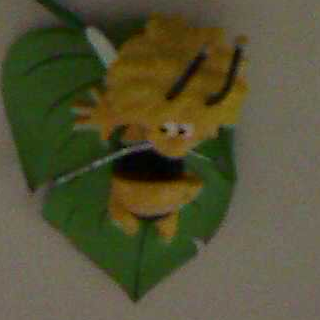} &
       \hspace{-3mm}
       \includegraphics[width=0.16\linewidth]{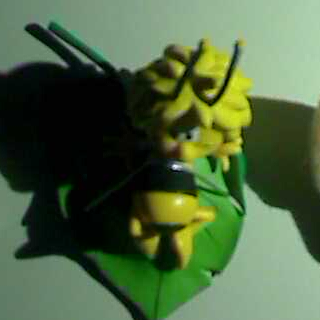} \\

       \includegraphics[width=0.16\linewidth]{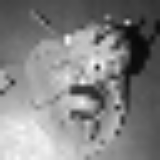} &
       \hspace{-3mm}
       \includegraphics[width=0.16\linewidth]{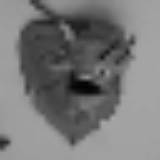} &
       \hspace{-3mm}
       \includegraphics[width=0.16\linewidth]{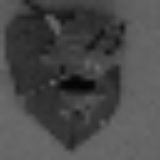} &
       \hspace{-3mm}
       \includegraphics[width=0.16\linewidth]{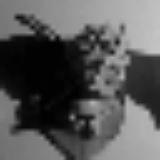} \\
       \includegraphics[width=0.16\linewidth]{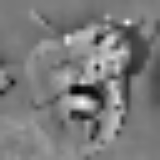} &
       \hspace{-3mm}
       \includegraphics[width=0.16\linewidth]{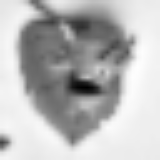} &
       \hspace{-3mm}
       \includegraphics[width=0.16\linewidth]{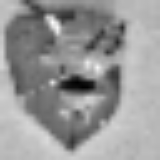} &
       \hspace{-3mm}
       \includegraphics[width=0.16\linewidth]{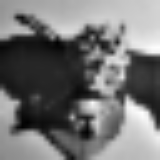} \\
       (a) & \hspace{-3mm} (b) & \hspace{-3mm} (c) & \hspace{-3mm}(d)  \\ 

     \end{tabular} & 
     \begin{tabular}{c}
       \includegraphics[trim=0.7cm 0.1cm 1.5cm 0.05cm,clip=true, width=0.25\linewidth]{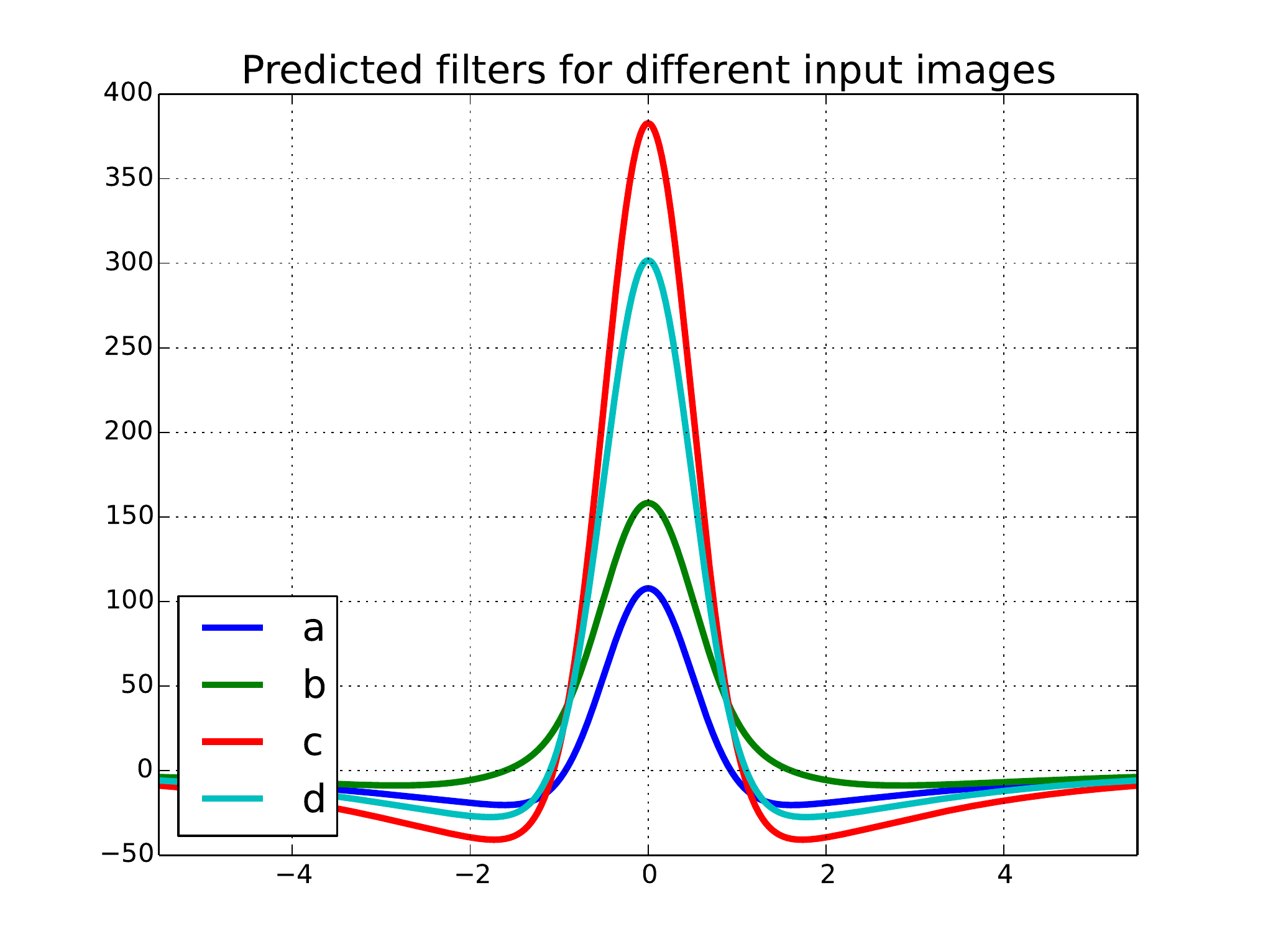}\\
       \includegraphics[trim=0.7cm 0.0cm 1.5cm 0.05cm,clip=true, width=0.25\linewidth]{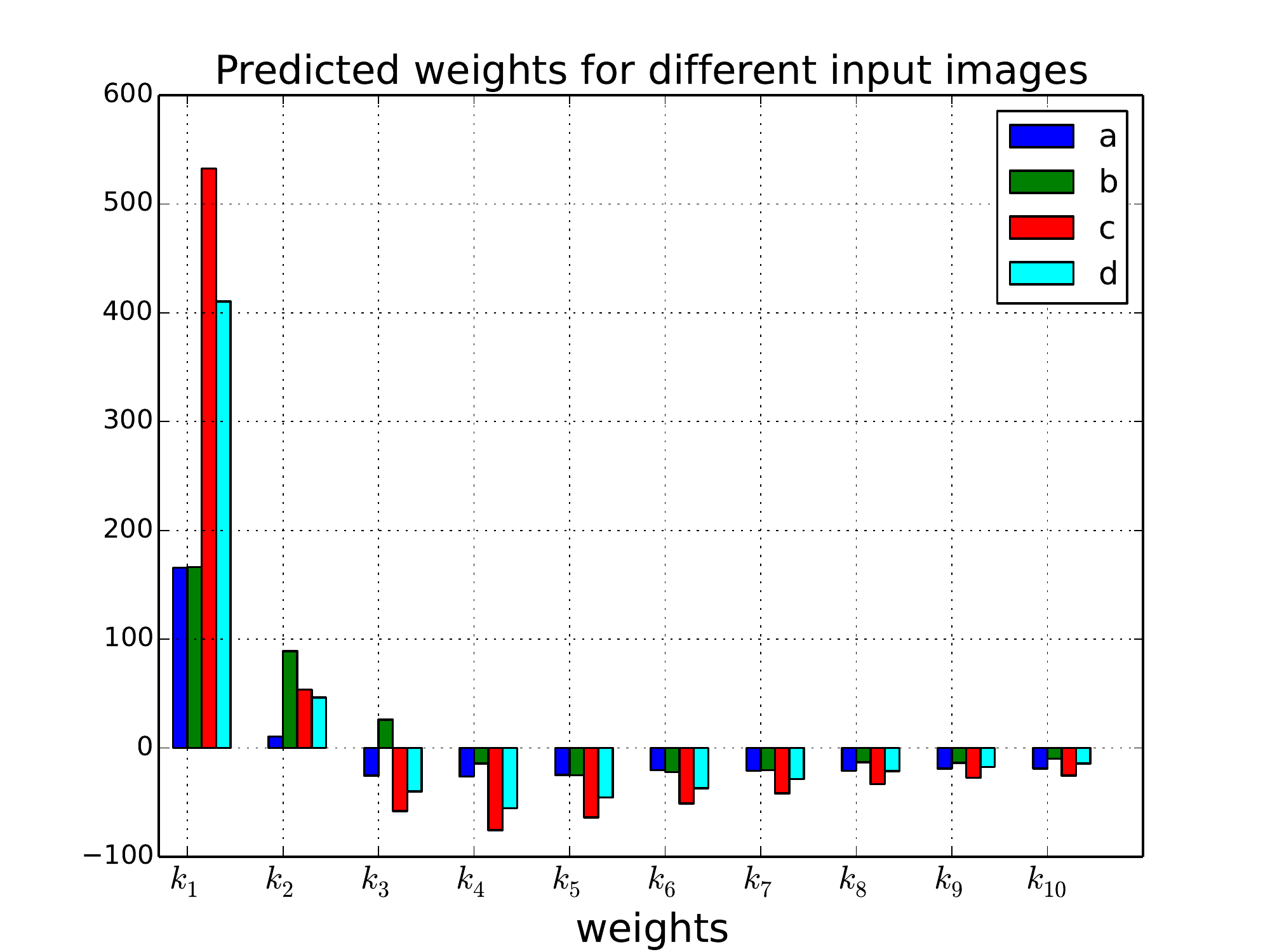}
     \end{tabular} \\
    
   \end{tabular}
 \end{center}
 \caption{Left  (a-d):  Four input  window  images  of  the plastic  toy  under
   different  illuminations. First  row:  original window.  Second row:  window
   after downscaling,  and used as input  to the normalizer. Third  row: window
   after normalization by  the filter predicted by  the Normalizer.  Top-right:       Predicted  filters using  the  model of  Eq.~\eqref{eq:AdaptiveLCN} for  the
   image windows a-d,  shown in 1D for clarity. The  filters predicted for dark
   images are larger  than the ones predicted for  bright images. Bottom-right:
   The 10 predicted coefficients $w_i$ for the image windows a-d.}
 \label{fig:gaussianFilters}
\end{figure*}

Fig.~\ref{fig:Retina_parameterization} shows that the optimized parameters found
for one  subset are only optimized  for that subset  and not the other  one.  It
also shows  that larger values  for $\sigma_1^\Ret$ and  $\sigma_2^\Ret$ perform
better on the  dark test images.  Fig.~\ref{fig:gaussianFilters}  shows that our
adaptive  Normalizer  learns  to   reproduce  this  behavior,  applying  larger
receptive fields to darker images and vice-versa.

\begin{figure*}[t]
  \begin{center}
    \begin{tabular}{cccc}

    \includegraphics[width=0.2\linewidth]{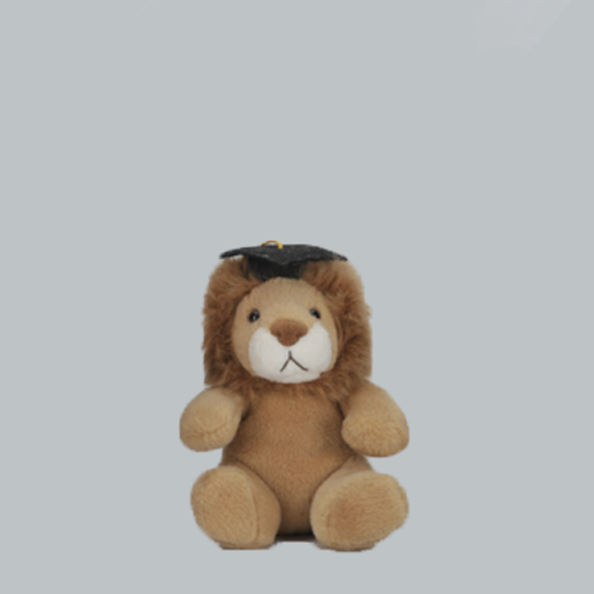} &
    \includegraphics[width=0.2\linewidth]{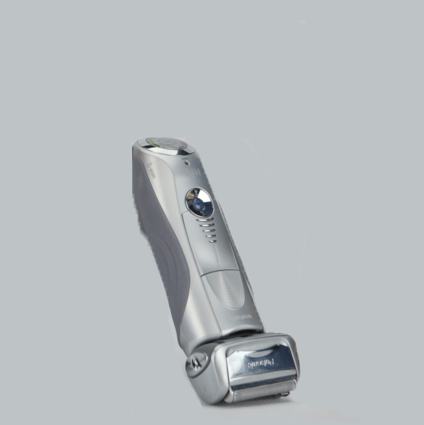} &
    \includegraphics[width=0.2\linewidth]{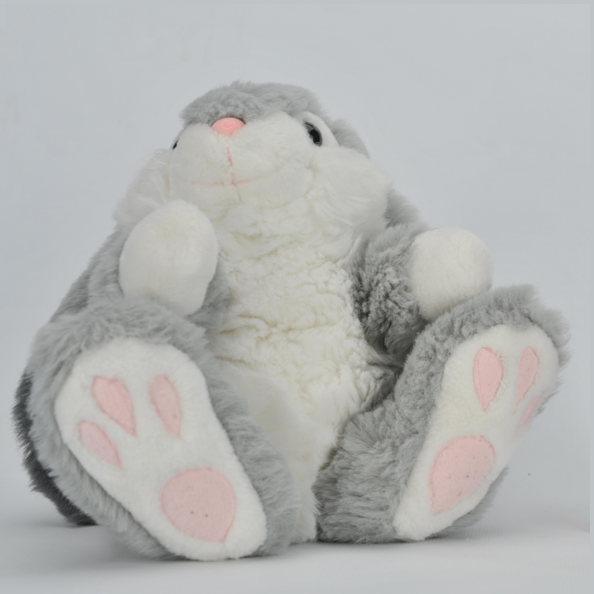} &
    \includegraphics[trim=1.0cm 0.5cm 0cm 1.35cm,clip=true,width=0.3\linewidth]{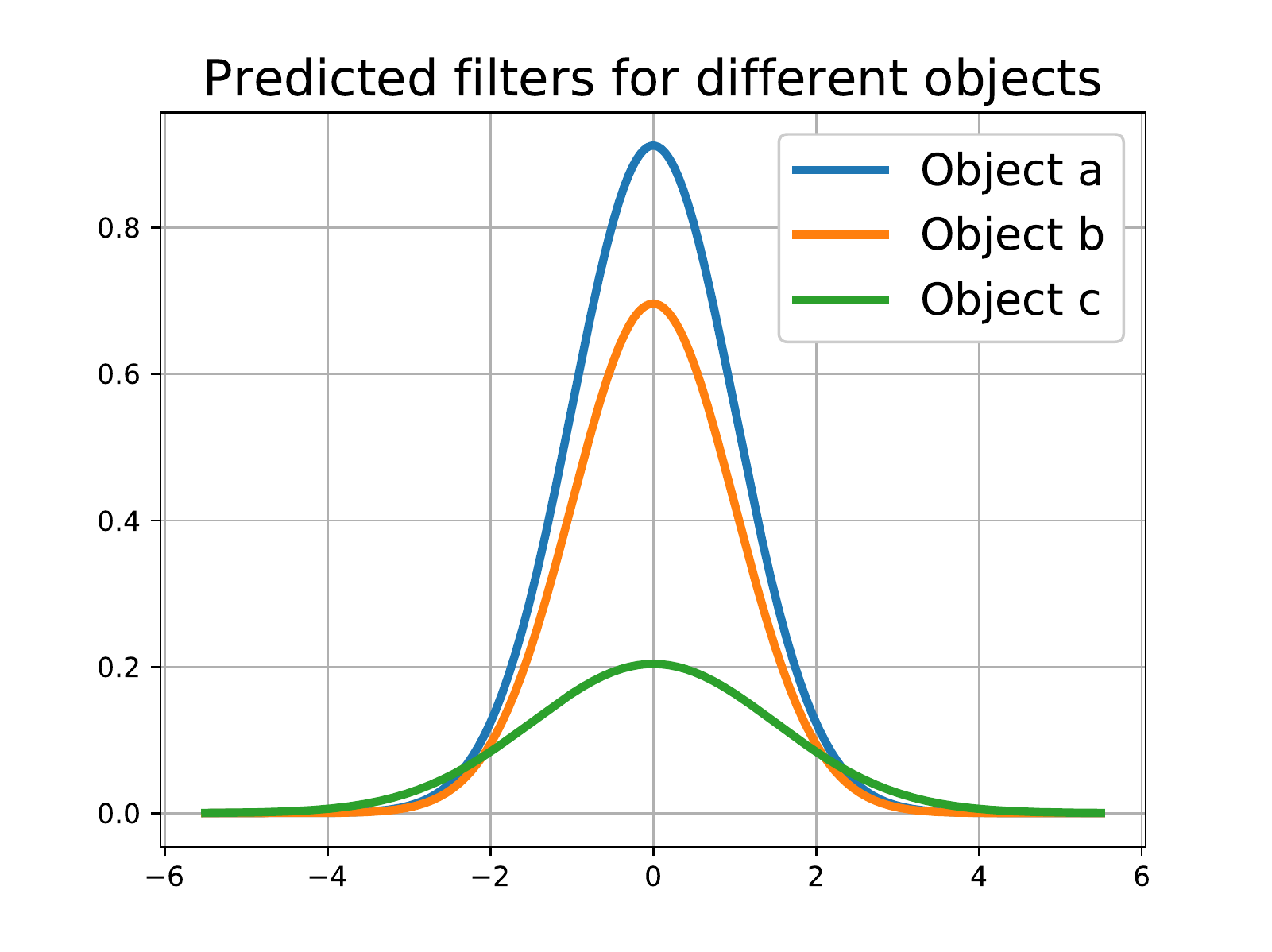}\\
    Object a & Object b & Object c & Predicted filters\\
    \end{tabular}
  \end{center}
  \caption{Predicted filters for DoG normalization trained on different objects.}
  \label{fig:alcn_params}
\end{figure*}

\paragraph{\bf{Different Objects}}

In order  to evaluate how different objects with different shapes and material affect on the predicted parameters,  we optimize on only one object of the Phos dataset at a time. As illustrated in Fig.~\ref{fig:alcn_params}, different kernels are learned for different objects. 


\subsection{From Window to Image Normalization}
\label{sec:imageNormalization}

Once trained on windows, we apply the Normalizer to the whole input
images by extending Eq.~\eqref{eq:AdaptiveLCN} to
\begin{equation}
  \ALCN({\bf I}) = \sum_{k=1}^N \;G_{\sigma_k^\ALCN} * \left(F_k({\bf I})\; \circ {\bf I}\right) \> ,
  \label{eq:AdaptiveLCNImage}
\end{equation}
where $F_k({\bf  I})$ is a  weight matrix with the  same dimension as  the input
image ${\bf I}$ for  the $k$-th 2D Gaussian filter, and  $\circ$ is the Hadamard
(element-wise) product.  The  weight matrix $F_k({\bf I})$  corresponding to the
$k$-th  2D Gaussian  filter  is computed  as  $\left(F_k({\bf I})\right)_{ij}  =
f_k(I_{ij})$, where  $(\cdot)_{ij}$ is the  entry in  the $i$-th row  and $j$-th
column of the matrix, $I_{ij}$ is the image window centered at $(i, j)$ in image
${\bf I}$, and $f_k(\cdot)$ is the $k$-th weight predicted by the Normalizer for
the  given image  window.  This  can  be done  very efficiently  by sharing  the
convolutions between windows~(\cite{Giusti13}).

Normalization is therefore different for each location of the input image.  This
allows  us to  adapt better  to the  local illumination  conditions. Because  it
relies  on Gaussian  filtering, it  is also  fast, taking only  50~ms for  10 2D
Gaussian filters, on an Intel Core  i7-5820K 3.30~GHz desktop with a GeForce GTX
980 Ti on a $128\times 128$ image.

\begin{figure*}
  \begin{center}
    \begin{tabular}{ccccc}
      \includegraphics[width=0.18\linewidth]{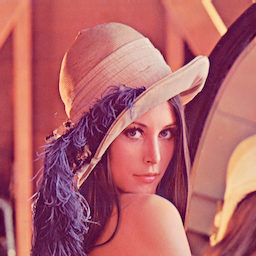} & 
      \includegraphics[width=0.18\linewidth]{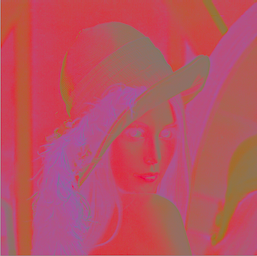} & 
      \includegraphics[width=0.18\linewidth]{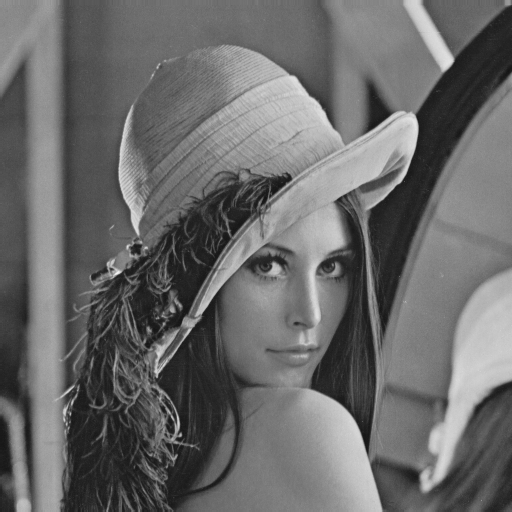} & 
      \includegraphics[width=0.18\linewidth]{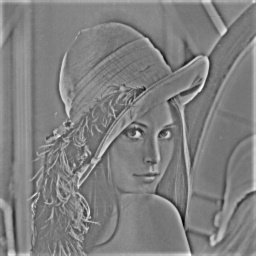} & 
      \includegraphics[width=0.18\linewidth]{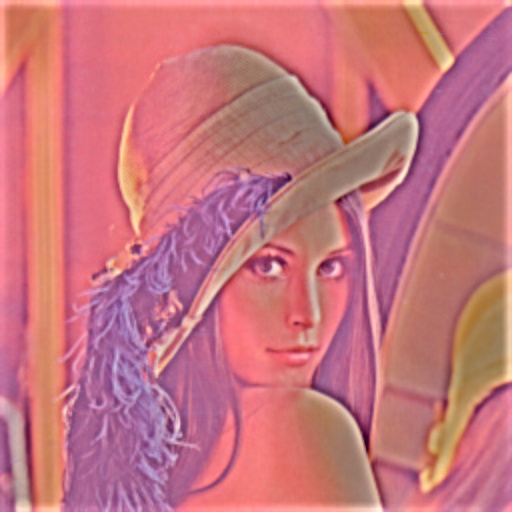} \\
      (a) & (b) & (c) & (d) & (e)\\
    \end{tabular}

  \end{center}
  \caption{\label{fig:normalized_images} (a): original RGB image. (b): ab
    channel in  CIE Lab  color space. (c):  grayscaled image.  (d): normalized
    grayscale image. (e): normalized color image.}
\end{figure*}

\subsection{Color Image Normalization}
\label{sec:colorImageNormalization}
For some  applications, such as 3D object pose estimation, it is important  to be
able to normalize not only grayscale images, but also color images as well, as
colors    bring    very    valuable     information.     To    do    so,    as
in~\cite{DBLP:journals/corr/ZhangIE16}, we first transform the input color image
in the CIE Lab colorspace, normalize the  lightness map L with our method, and
re-transform the image in the RGB  space without changing the ab channels.  An
example  of  a   normalized  color  image  using  this  method   is  shown  in
Fig.~\ref{fig:normalized_images}(e).

\subsection{Network Architecture and Optimization Details}
\label{sec:architecture}
A  relatively simple  architecture is  sufficient for  the Normalizer:  In all of our
experiments, the  first layer performs  20 convolutions with   $5 \times 5$  filters
with  $2 \times 2$   max-pooling.   The   second  layer  performs   50   $5 \times 5$ 
convolutions followed  by $2 \times 2$  max-pooling.  The third  layer is  a fully
connected layer  of 1024  hidden units.   The last  layer returns  the predicted
weights. In order to keep optimization tractable,  we downscaled the training  images of
the target objects by  a factor of 10.  To avoid border effects,  we use $48 \times
48$ input patches for the Normalizer, and use $32 \times 32$ patches as input to
the  Detector.  We  use  the  $\tanh$ function  as  activation  function, as  it
performs better than  ReLU on our problem. This difference  is because a sigmoid
can better control  the large range of intensities exhibited in the images of our
dataset, while other datasets have much more controlled illuminations.

\subsection{Generating Synthetic Images}
\label{sec:gen_synch_img}
Since the Phos dataset is small, we augment it by applying simple
random transformations to the original images.
We segmented the objects manually, so that we can change the
background easily.

We  experimented with  several methods to artificially change the illuminations,  and we
settled to the following formulas to generate a new image $\imagenew$
given an original image $\imageref$.
We first scale the original image, randomly replace the background, and scale
the pixel intensities:
\begin{equation}
  \imageint = a (  \bg( \scale_s(\imageref ) ) + b \> ,
\end{equation}
where $a$, $b$, and $s$ are value randomly sampled from the ranges $[1-A; 1+A]$,
$[-B;  +B]$, and  $[1-S; 1+S]$  respectively.  $\bg(\cdot)$  is a  function that
replaces the background of  the image by a random background, which  can be uniform or
cropped    from   an    image   from    the   ImageNet    (dataset~\cite{Deng09}).
$\scale_s(\cdot)$ is a  function that upscales or downscales  the original image
by a factor $s$.  

The generated image is then taken as
\begin{equation}
  \imagenew = \clip( G(\imageint)) \> ,
\label{eq:imagenew}
\end{equation}
where $G(\cdot)$  adds Gaussian noise,  and $\clip(\cdot)$ is the  function that
clips the intensity  values to the $[0;255]$ interval.  This  function allows us
to simulate saturation effects, and makes the transformation non-linear, even in
the absence  of noise. $\imageint$ is  an intermediate image that  can influence
the amount of  noise: In all our  experiments, we use $A = 0.5$,  $B = 0.4$,
and $S = 0.1$.

We generate 500,000 synthetic images, with the same number of
false and negative images.  Once the  Normalizer is trained on the Phos dataset,
we can use synthetic  images created from a very small number  of real images of
the target objects to train a new classifier to recognize these objects: Some of
our experiments presented  below use only one  real image. At test  time, we run
the  Detector on  all  $48 \times 48$ image  windows extracted  from the  test
image.


\section{Experiments}

\label{sec:results}
\label{sec:experiments}

In this  section, we  first introduce  the datasets we  used for  evaluating the
methods described in  the previous section, including our own.   We then present
the network architecture and optimization details.  We perform thorough ablation
studies to demonstrate our contribution,  by benchmarking on 2D object detection
under  drastic illumination  changes  when  only few  images  are available  for
training.  We  finally evaluate our  normalization to improve the  robustness to
light changes of state-of-the-art 3D  object detection and pose estimation, face
recognition and semantic segmentation methods.

\newcommand{\myheight}{0.15\linewidth}
\newcommand{\Objectheight}{0.13\linewidth}
\begin{figure*}[t]
  \begin{center}
    \begin{tabular}{cccc}

    \begin{picture}(65,80)
      \put(0,10){\includegraphics[trim=16cm 8.5cm 17cm 7cm,clip=true,height=\Objectheight]{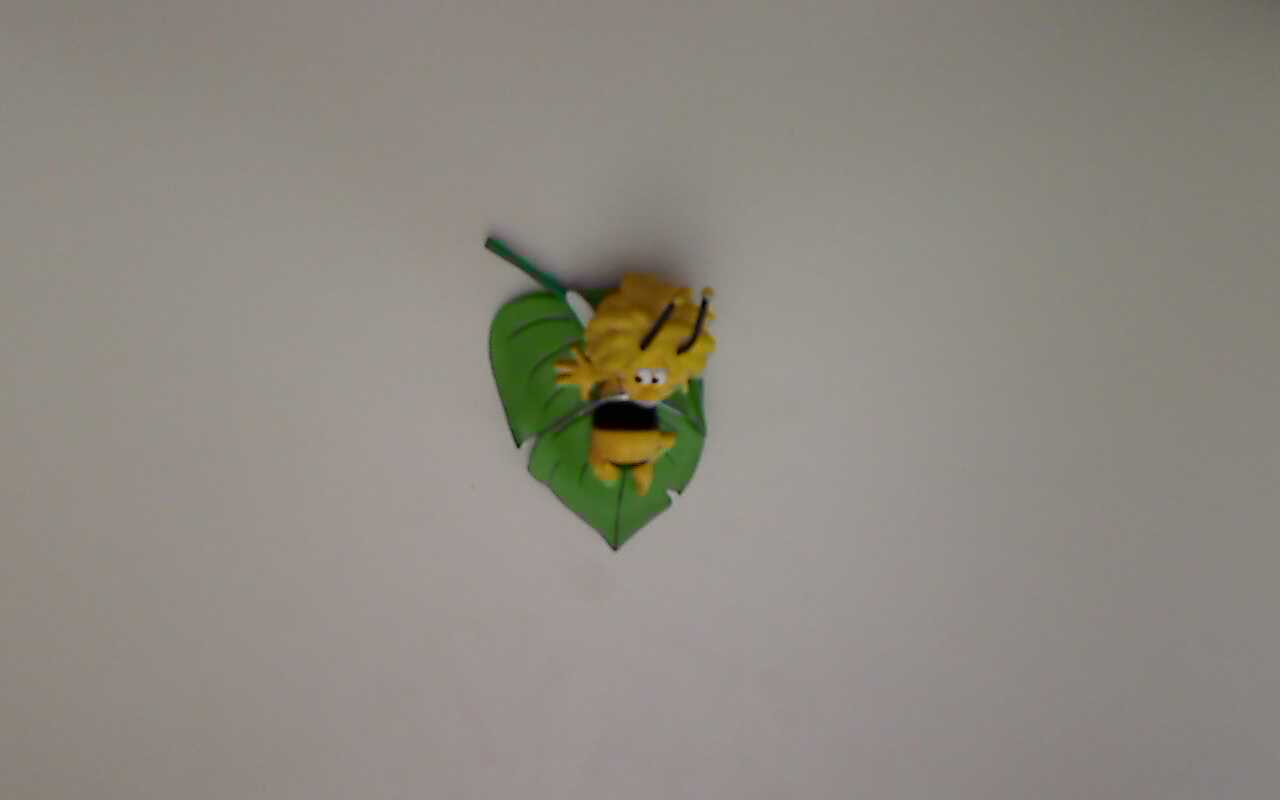}}
      \put(12,0){ Object \#1}
    \end{picture} & 
    \includegraphics[height=\myheight]{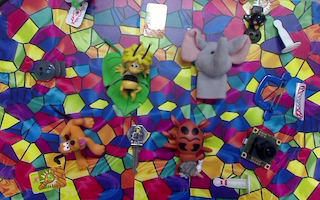}&
    \includegraphics[height=\myheight]{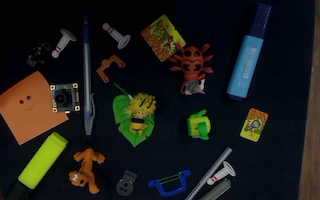}&
    \includegraphics[height=\myheight]{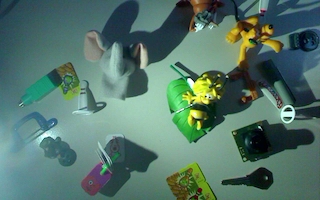}\\

    \begin{picture}(65,80)
      \put(0,10){\includegraphics[trim=16cm 8.5cm 17cm 7cm,clip=true,height=\Objectheight]{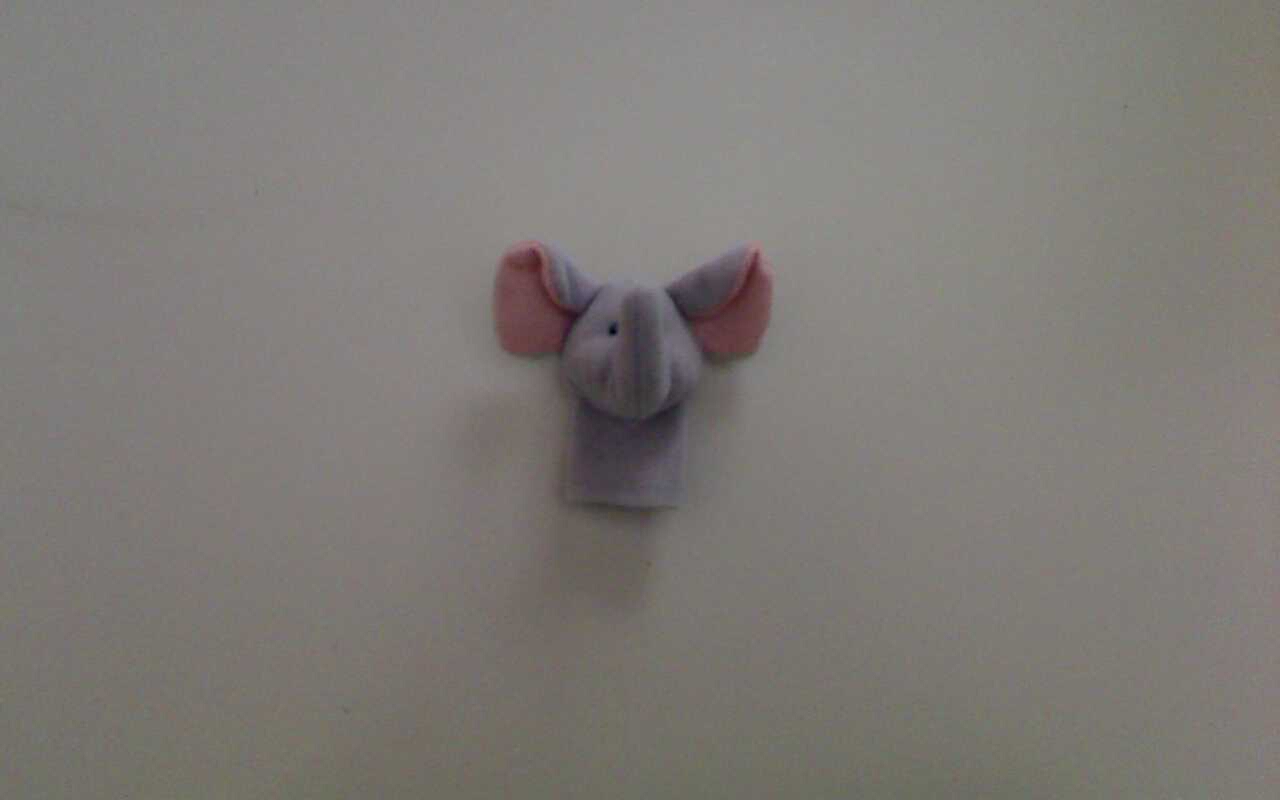}}
      \put(12,0){ Object \#2}
    \end{picture} & 
    \includegraphics[height=\myheight]{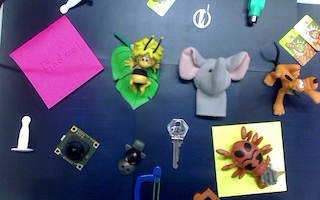}&
    \includegraphics[height=\myheight]{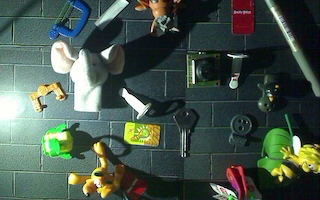}&
    \includegraphics[height=\myheight]{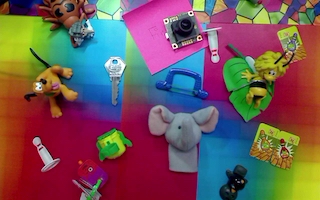}\\

    \begin{picture}(65,80)
      \put(0,10){\includegraphics[trim=16cm 8.5cm 17cm 7cm,clip=true,height=\Objectheight]{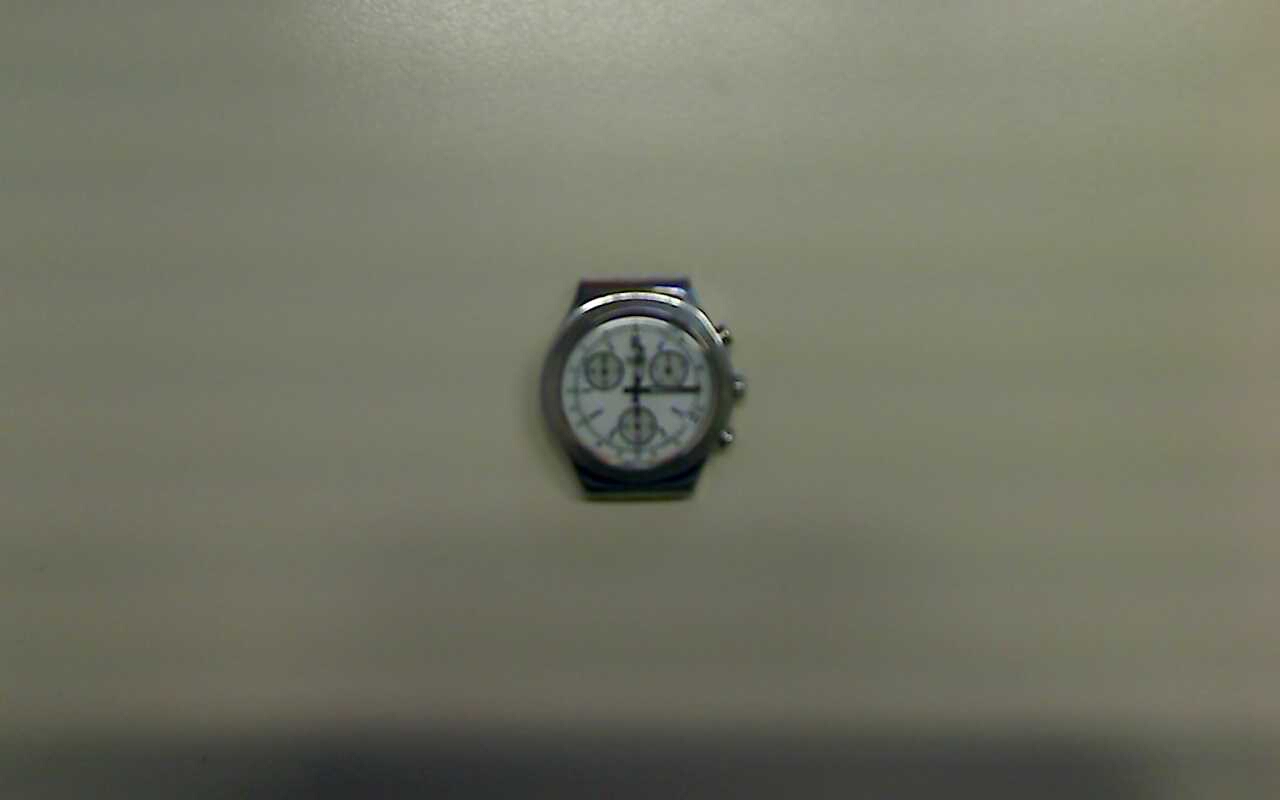}}
      \put(12,0){ Object \#3}
    \end{picture} & 
      \includegraphics[height=\myheight]{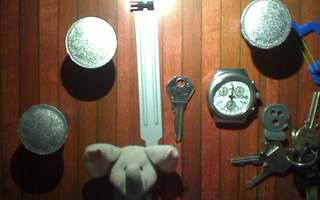}&
      \includegraphics[height=\myheight]{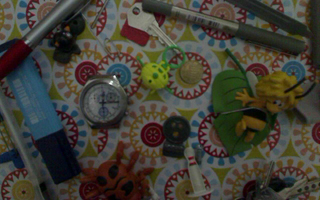}&
      \includegraphics[height=\myheight]{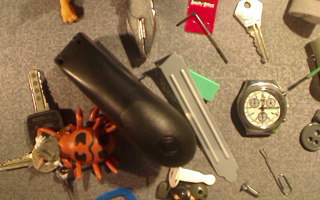}\\
    \end{tabular}
  \end{center}
  \caption{The objects for our ALCN-2D  dataset, and representative test images.  We
  selected  three objects  spanning  different  material properties: plastic (Object \#1),
  velvet (Object \#2), metal (Object \#3)  (velvet has a BRDF  that is neither Lambertian  nor specular,
  and the  metallic object---the watch---is  very specular).  By contrast with previous datasets, we
  have a  very large number of  test images (1200 for  each object), capturing
  many different  illuminations and background.}
  \label{fig:alcn_dataset}
\end{figure*}
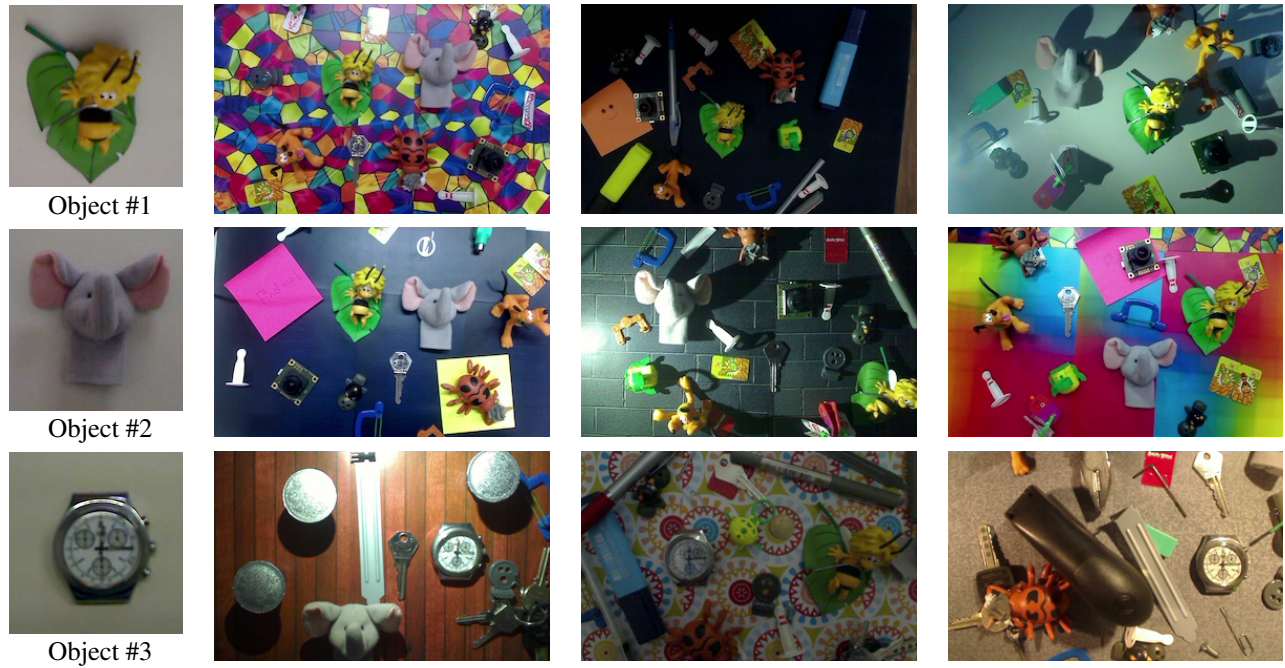


\subsection{Datasets}
\label{sec:datasets}

Some datasets  have instances  captured under  different illuminations,  such as
NORB~(\cite{LeCun04}), ALOI~(\cite{Geusebroek05}),  CMU Multi-PIE~(\cite{Gross09}) or
Phos~(\cite{phos2013}).  However, they are not suitable for our purposes: NORB has
only 6 different lighting directions; the images of ALOI contain a single object
only and over  a black background; CMU Multi-PIE
was developed for face recognition and the image is always centered on the face;
Phos was useful for our joint training approach, however, it has only 15 test
images and the objects are always at the same locations, which would make the
evaluation dubious.

We thus created  a new  dataset for  benchmarking object  detection under
challenging lighting conditions and cluttered background.  We will refer to this
dataset  as the  ALCN-2D dataset.   As shown  in Fig.~\ref{fig:alcn_dataset},  
we
selected three  objects spanning different material  properties: plastic (Object \#1), velvet (Object \#2)
and   metal  (Object \#3) (velvet   has   a    BRDF   that   is   neither   Lambertian   nor
specular~(\cite{Lu98}), and the metallic object---the watch---is very specular).
For each object, we have 10  $300 \times 300$ grayscale training images and
1200 $1280  \times 800$  grayscale test images,  exhibiting these  objects under
different  illuminations,  different lighting  colors,  and  distractors in  the
background. The number of test images is therefore much larger than for previous
datasets. We  manually annotated  the ground  truth bounding  boxes in  the test
images in which the target object is present.
In this first dataset, the objects  are intentionally moved on a planar surface,
in  order  to  limit  the  perspective  appearance  changes  and  focus  on  the
illumination variations.

The second dataset we consider is the BOX Dataset from
the  authors  of  \cite{Crivellaro15},  which  combines  perspective  and  light
changes.  It is made of a registered  training sequence of an electric box under
various 3D poses but a single illumination  and a test sequence of the same box
under various  3D poses and illuminations. Some images are shown  in the second
row  of  Fig.~\ref{fig:teaser}. This test  sequence  was  not actually part  of  the
experiments performed by  \cite{Crivellaro15} since it was  too challenging in scope. The goal  is to
estimate the 3D pose of the box.

Finally, we  introduce another dataset for  3D pose estimation. This  dataset is
made of  a training  sequence of  1000 registered  frames of  the Duck  from the
Hinterstoisser dataset~(\cite{Hinterstoisser12})  obtained by 3D printing  under a
single illumination and 8 testing  sequences under various illuminations.  Some
images are shown in the third row of Fig.~\ref{fig:teaser}. We will   refer to 
this dataset as the ALCN-Duck dataset.

\subsection{Experiments and Discussion}

For evaluation, we use the PASCAL criterion  to decide if a detection is correct
with an Intersection over Union of $0.8$, with fixed box sizes of 
$300 \times 300$,
reporting Precision-Recall~(PR) curves  and Areas Under Curve~(AUC) in order
to compare the performances of the different methods.

\subsubsection{Explicit Normalization vs Illumination Robustness with Deep Learning}

As  mentioned  in  the  introduction,  Deep Networks  can  learn  robustness  to
illumination  variations without  explicitly  handling them,  at  least to  some
extent.  To  show that our  method allows  us to go  further, we first  tried to
train several Deep Network architectures from scratch,  without normalizing the
images beforehand, by varying the number of layers and the number of filters for
each layer.   We use  one real  example of each  object in  the ALCN-2D
  dataset for this experiment. The best  architecture we found performs with an
AUC of 0.606.  Our method, however, still performs better with  an AUC of 0.787.
This shows that  our approach achieves better robustness to  illumination than a
single CNN, at least when the training set is limited, as in our scenario.

We      also       evaluated      Deep      Residual       Learning      Network
architectures~(\cite{he2016deep}).   We used  the  same  network architectures  and
training parameters as  in \cite{he2016deep} on CIFAR-10.  ResNets  with 20, 32,
44, 56 and 110 layers perform with  AUCs of 0.456, 0.498, 0.518, 0.589 and 0.565
respectively, which  is still outperformed  by a  much simpler network  when our
normalization  is  used.   Between  56   and  110  layers,  the  network  starts
overfitting,  and increasing  the  number of  layers results  in  a decrease  of
performance.


\subsubsection{Comparing ALCN against previous Normalization Methods}
\label{sec:alcn_vs_normalization_methods}
\begin{figure*}
  \begin{center}
    \begin{tabular}{ccc}
      \includegraphics[trim=0.7cm 0.1cm 1.5cm 0.05cm,clip=true, 
      width=0.30\linewidth]{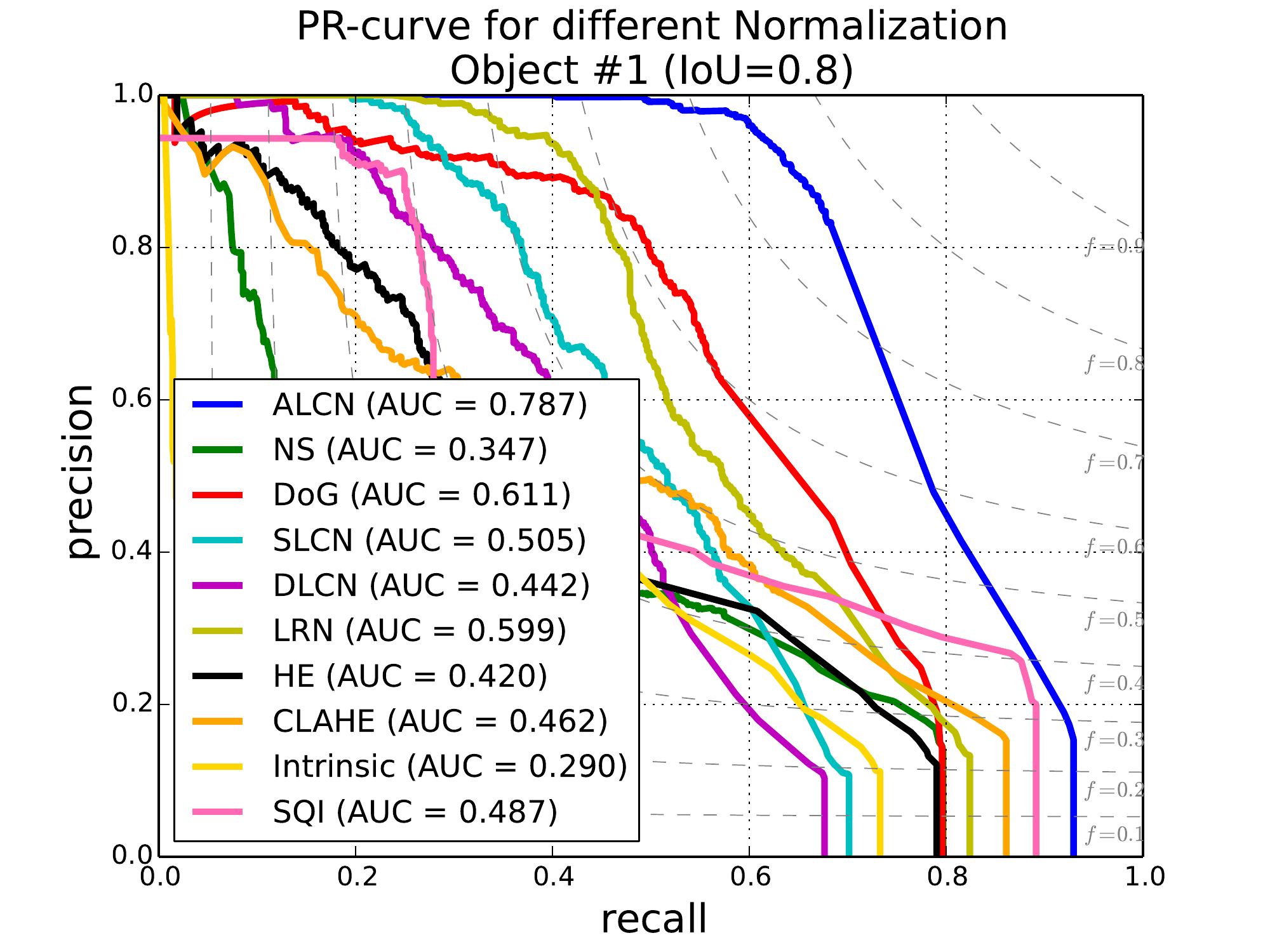} &
      \includegraphics[trim=0.7cm 0.1cm 1.5cm 0.05cm,clip=true, 
      width=0.30\linewidth]{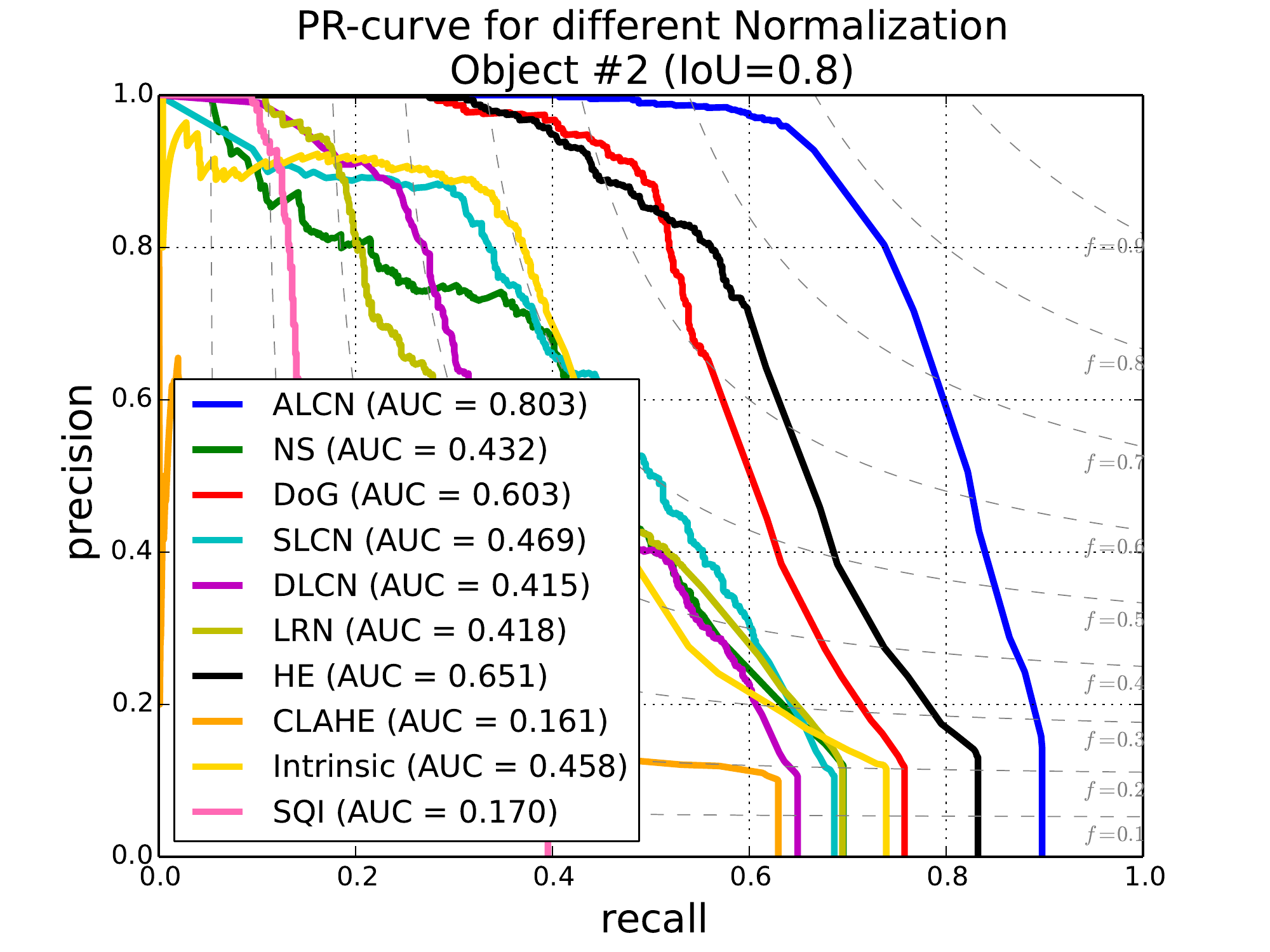} &
      \includegraphics[trim=0.7cm 0.1cm 1.5cm 0.05cm,clip=true, 
      width=0.30\linewidth]{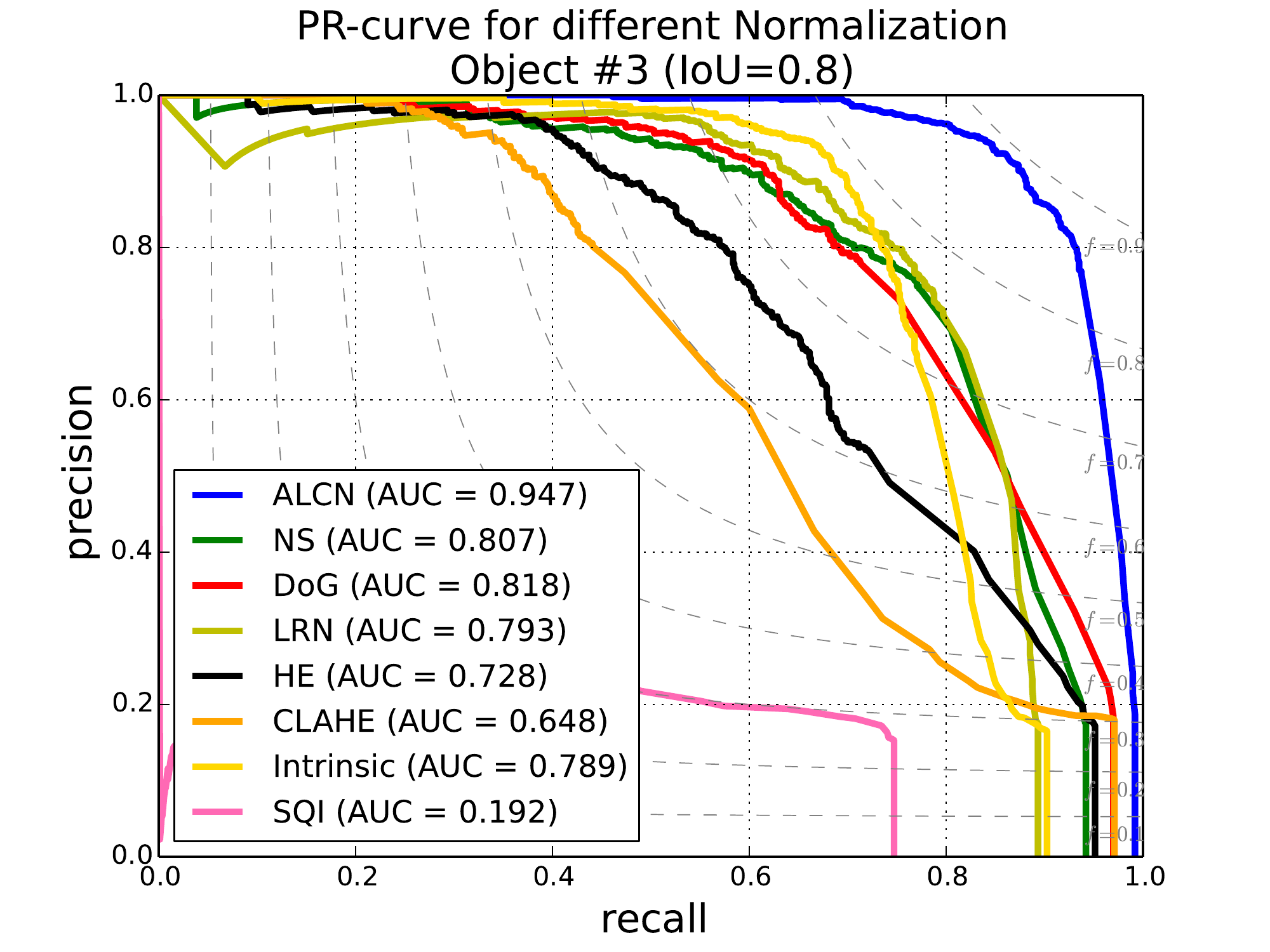} \\
    \end{tabular}
  \end{center}
  \caption{\label{fig:gn-sub-div}
    Comparing  different normalization  methods  using  the best  parameter
    values for each method for Objects~\#1,~\#2 and~\#3 of ALCN-2D.  ALCN 
    systematically performs best by
    a large margin. }
\end{figure*}

In our evaluations, we consider different existing methods described in Section~\ref{sec:overview_existing_normalizations}.
In order  to assess  the effects  of different  normalization techniques  on the
detection performances, we  employed the same detector architecture  for the
normalization  methods,  but  re-training  it for  every  normalization  method.
Fig.~\ref{fig:gn-sub-div} compares these methods on the ALCN-2D dataset.  For
DoG, Subtractive and Divisive LCN, we optimized their parameters to perform best
on the training set.  We tried different method of intrinsic image decomposition and they perform with similar accuracy. In this paper, we use the implementation of (\cite{shen2013intrinsic}), which performs slightly better on the ALCN-2D dataset, compare to other implementations. Our method  consistently outperforms the others for all 
objects of  the ALCN  dataset.  Most  of the other  methods have  very different
performances across the different objects of the dataset.  Whitening obtained an
extremely  bad score  for all  objects,  while both  versions of  LCN failed  in
detecting Object~\#3, the  most specular object, obtaining an  AUC score lower
than 0.1.

\begin{figure*}
  \begin{center}
    \begin{tabular}{ccc}
      \includegraphics[trim=0.7cm 0.1cm 1.5cm 0.05cm,clip=true, 
      width=0.30\linewidth]{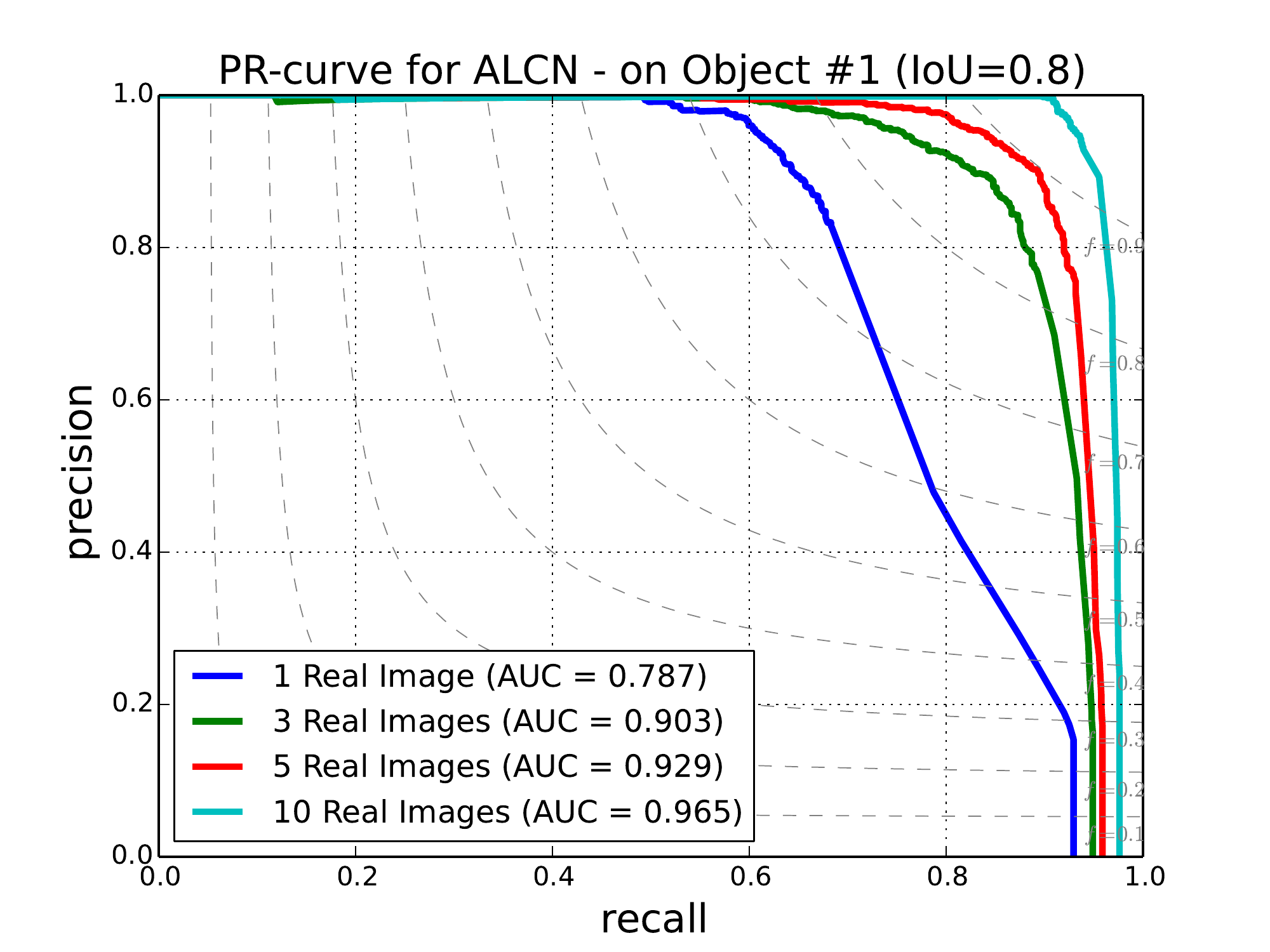}  &
      \includegraphics[trim=0.7cm 0.1cm 1.5cm 0.05cm,clip=true, 
      width=0.30\linewidth]{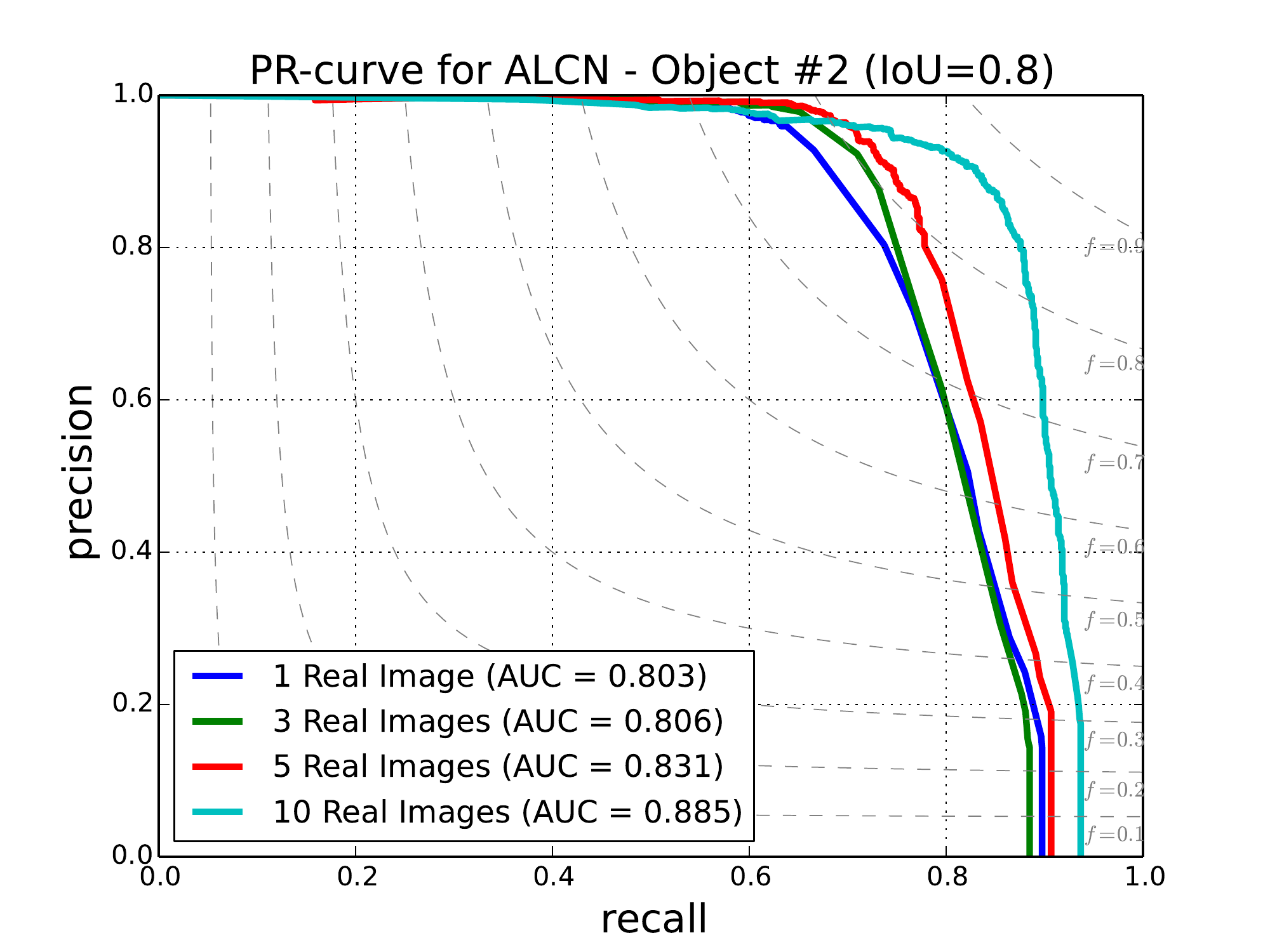}  &
      \includegraphics[trim=0.7cm 0.1cm 1.5cm 0.05cm,clip=true, 
      width=0.30\linewidth]{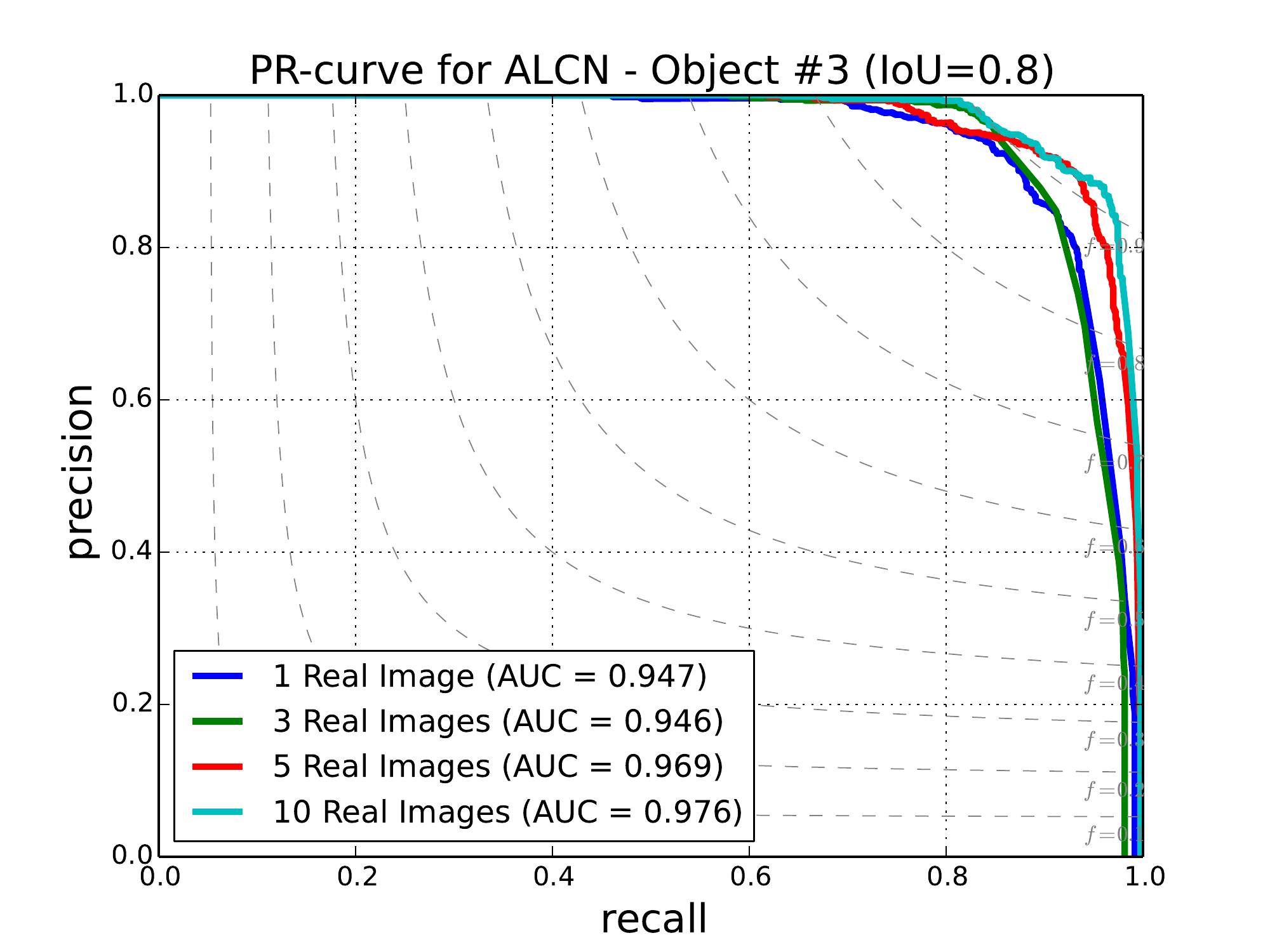}  \\
    \end{tabular}
  \end{center}
  \caption{\label{fig:nb_samples}
    Evaluating the influence of the  number of real images
    used for training the detector  on Objects~\#1,~\#2 and~\#3 of ALCN-2D.  The 
    detection accuracy keeps
    increasing when using more real images for generating the training set.}
\end{figure*}


\subsubsection{Impact of Number of Real Images}
\begin{figure}
  \begin{center}
    \includegraphics[trim=2.52cm 0.1cm 0.05cm 0.85cm,clip=true,width=0.95\linewidth]{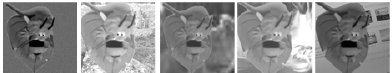}
  \end{center}
  \caption{Some synthetic  images generated  from the  object \#1  shown in
    Fig.~\ref{fig:alcn_dataset}.}
  \label{fig:synth_generated_images}
\end{figure}

Once the Normalizer is trained on the Phos dataset, we freeze its weights and plug it to a detector to detect target object. To train the Detector, we use 500,000 synthetically generated images with the same way as described in Section~\ref{sec:gen_synch_img}. Some synthetic  images generated are shown in Fig.~\ref{fig:synth_generated_images}. These 500,000 images can be generated either from only one single real image, or more. In Section~\ref{sec:alcn_vs_normalization_methods}, we showed that using only
one real image to generate the whole training set already gives us good results. Fig.~\ref{fig:nb_samples} illustrates that using
more real images while keeping the total number of synthetically generated images same as before,  improves the performances further, 10 real images are  enough for very good
performance. This shows that we can learn to detect objects under very different
drastic  illuminations  from  very  few  real  examples  augmented  with  simple
synthetic examples.

 \subsubsection{Activation Functions}
 \label{sec:activationFunction}
 While  sigmoid functions  were originally  used in  early neural  networks and
 CNNs, the  popular choice  is now  the ReLU operator,  because it  often eases
 tuning the convergence as the derivatives  are constant, while special care is
 to be taken when using sigmoids.

\begin{figure}
   \begin{center}
     \includegraphics[width=0.90\linewidth]{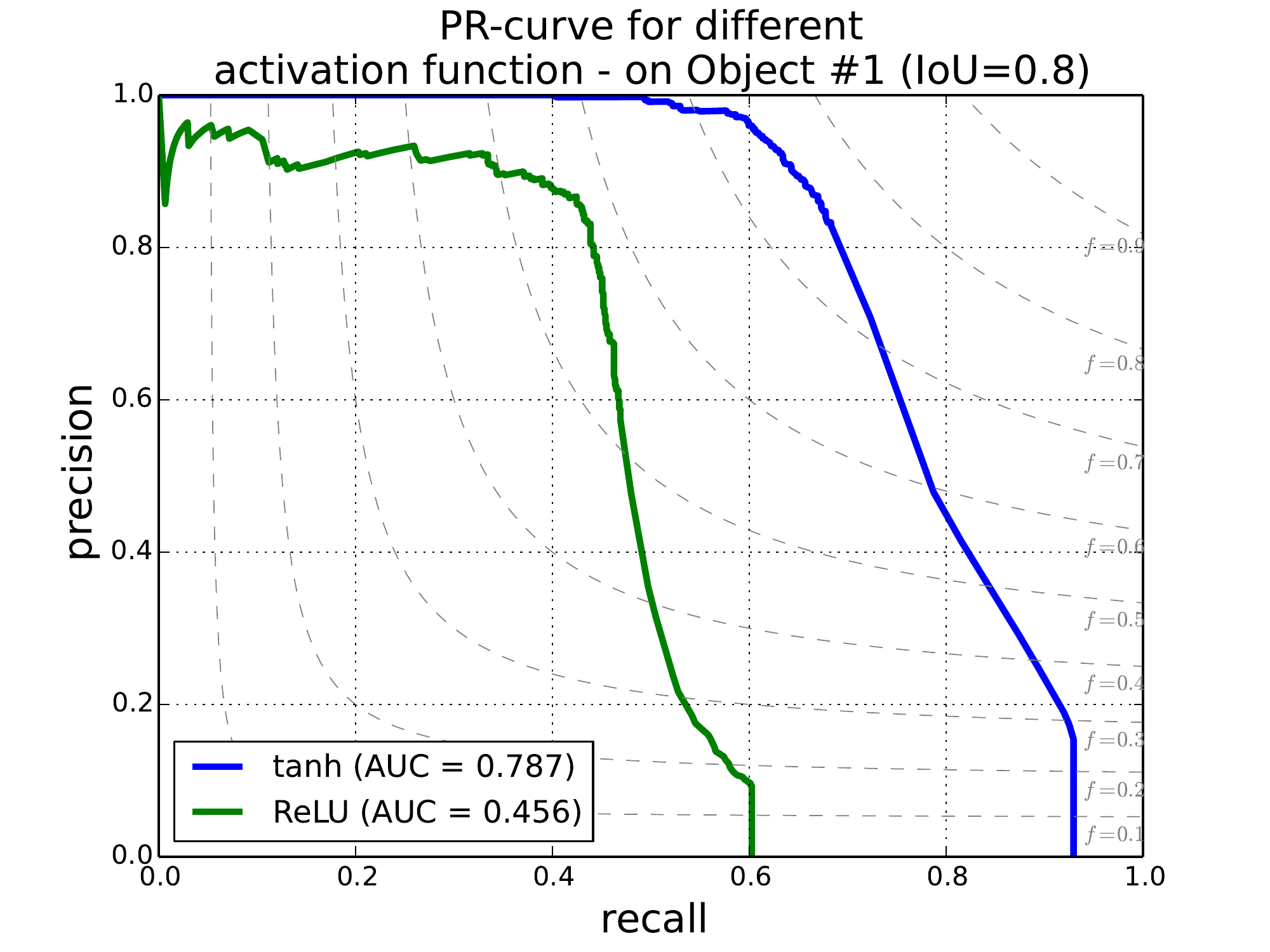}
   \end{center}
   \vspace{-0.27cm}
   \caption{\label{fig:relu-vs-tanh} Influence  of the activation  function.  The
     plots show the  PR curves using our normalization, applying either
     the ReLU  operator, or the  sigmoid $\tanh$ function as  activation function.
     $\tanh$  appears to perform better, probably because
     it helps to control the range of intensity values in our test images.}
 \end{figure}

 However, Fig.~\ref{fig:relu-vs-tanh}  shows that using the  hyperbolic tangent
 $\tanh$ sigmoid  function yields  clearly better results  than using  the ReLU
 activation functions on  our problem.  This  difference is  because a sigmoid  can control
 better the large range of intensities  exhibited in the images of our dataset,
 while other datasets have much more controlled illuminations.

\begin{figure}
  \begin{center}
    \begin{tabular}{cc}
    \includegraphics[trim=0.7cm 0.1cm 1.5cm 0.85cm,clip=true, width=0.45\linewidth]{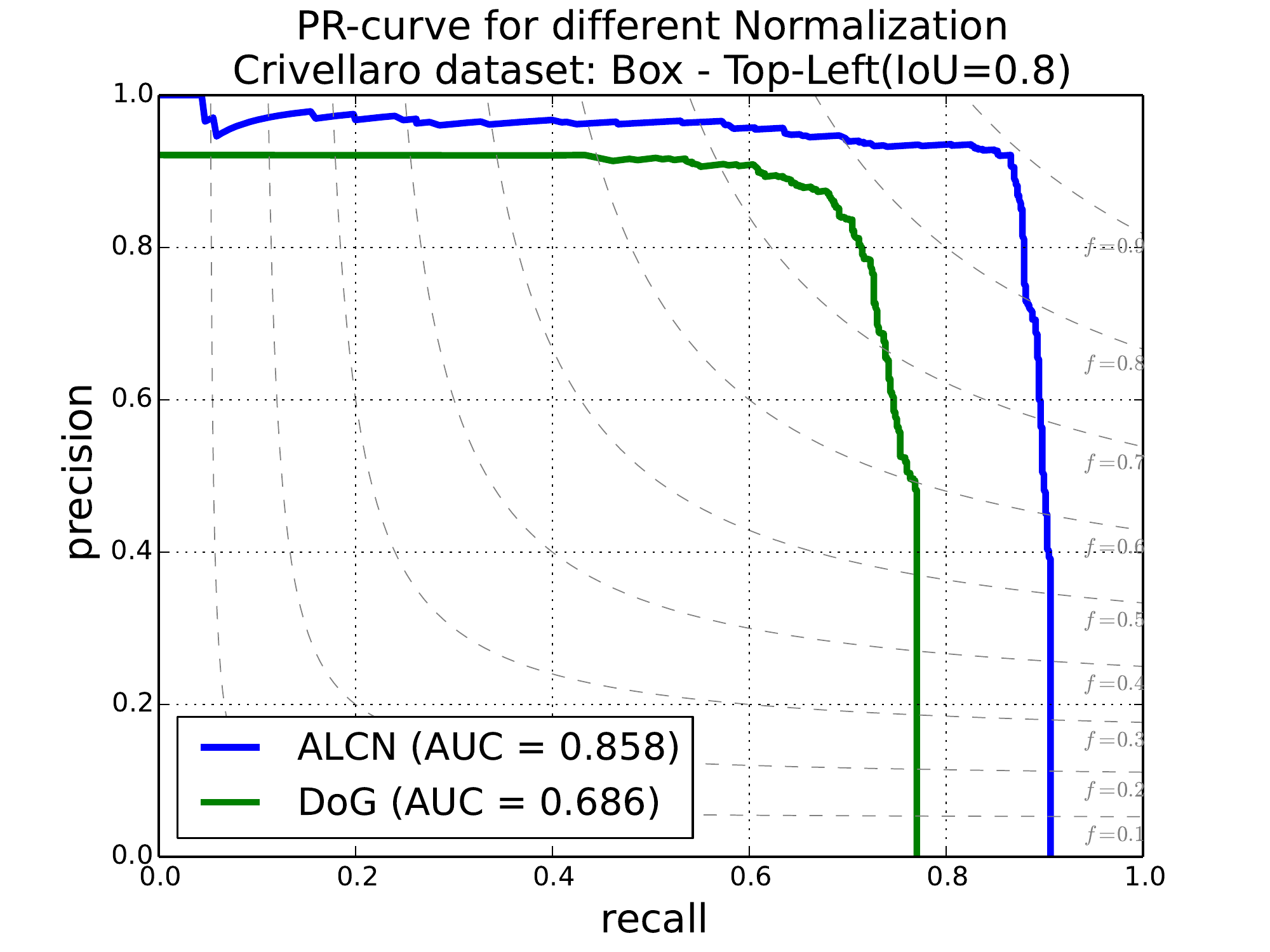} &
      \includegraphics[trim=0.7cm 0.1cm 1.5cm 0.85cm,clip=true, width=0.45\linewidth]{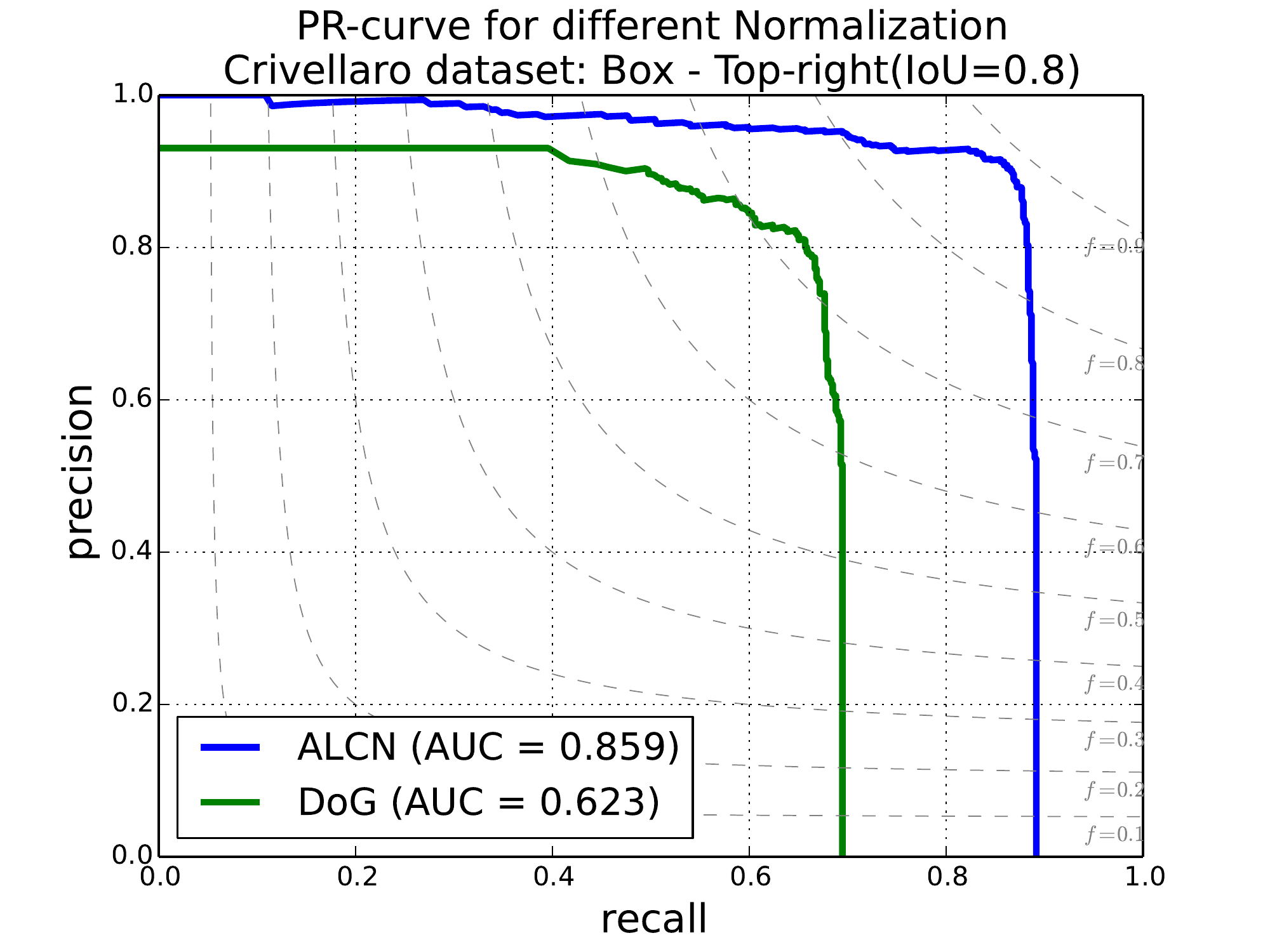} \\
    \includegraphics[trim=0.7cm 0.1cm 1.5cm 0.85cm,clip=true, width=0.45\linewidth]{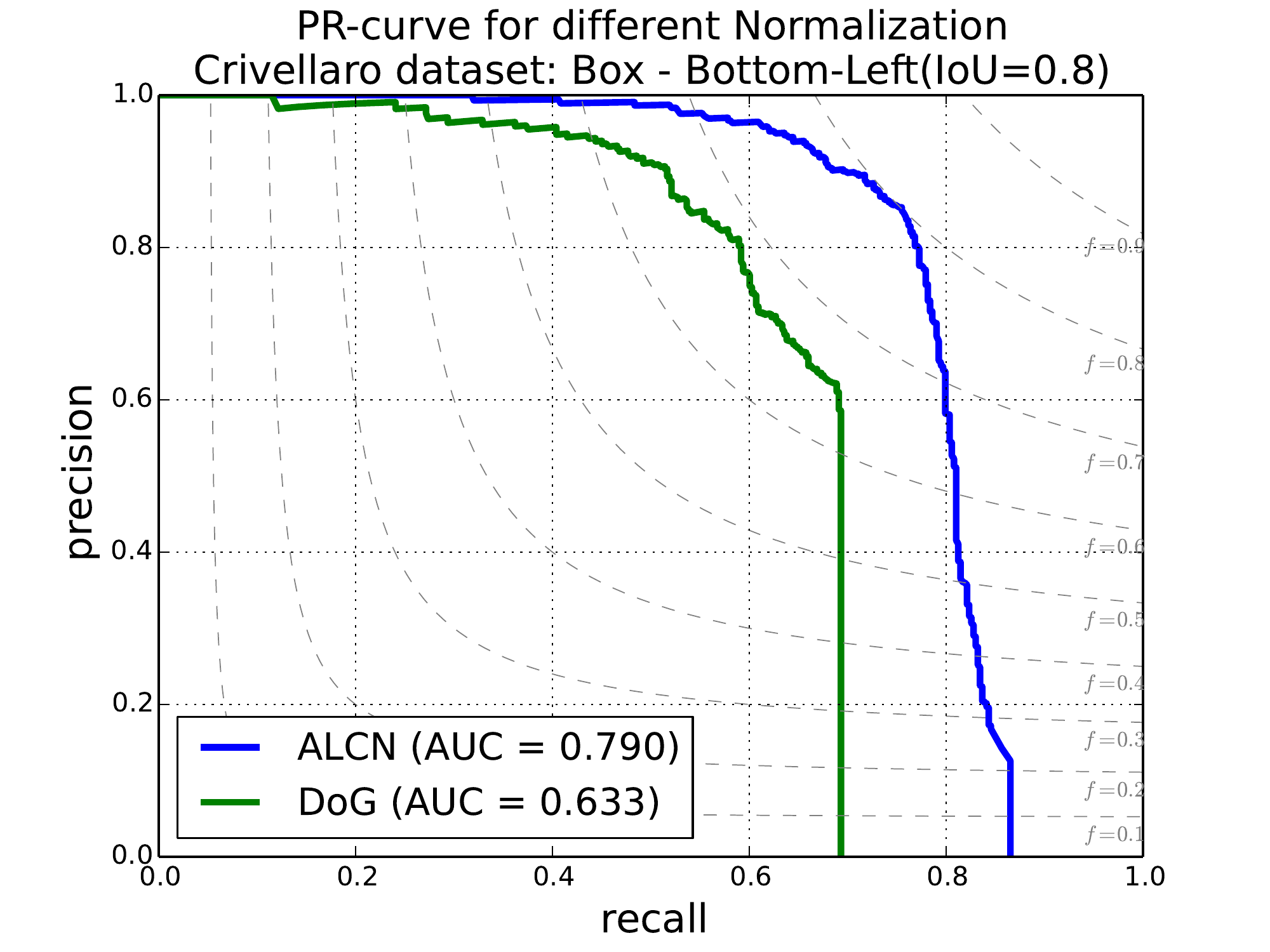} &
    \includegraphics[trim=0.7cm 0.1cm 1.5cm 0.85cm,clip=true, width=0.45\linewidth]{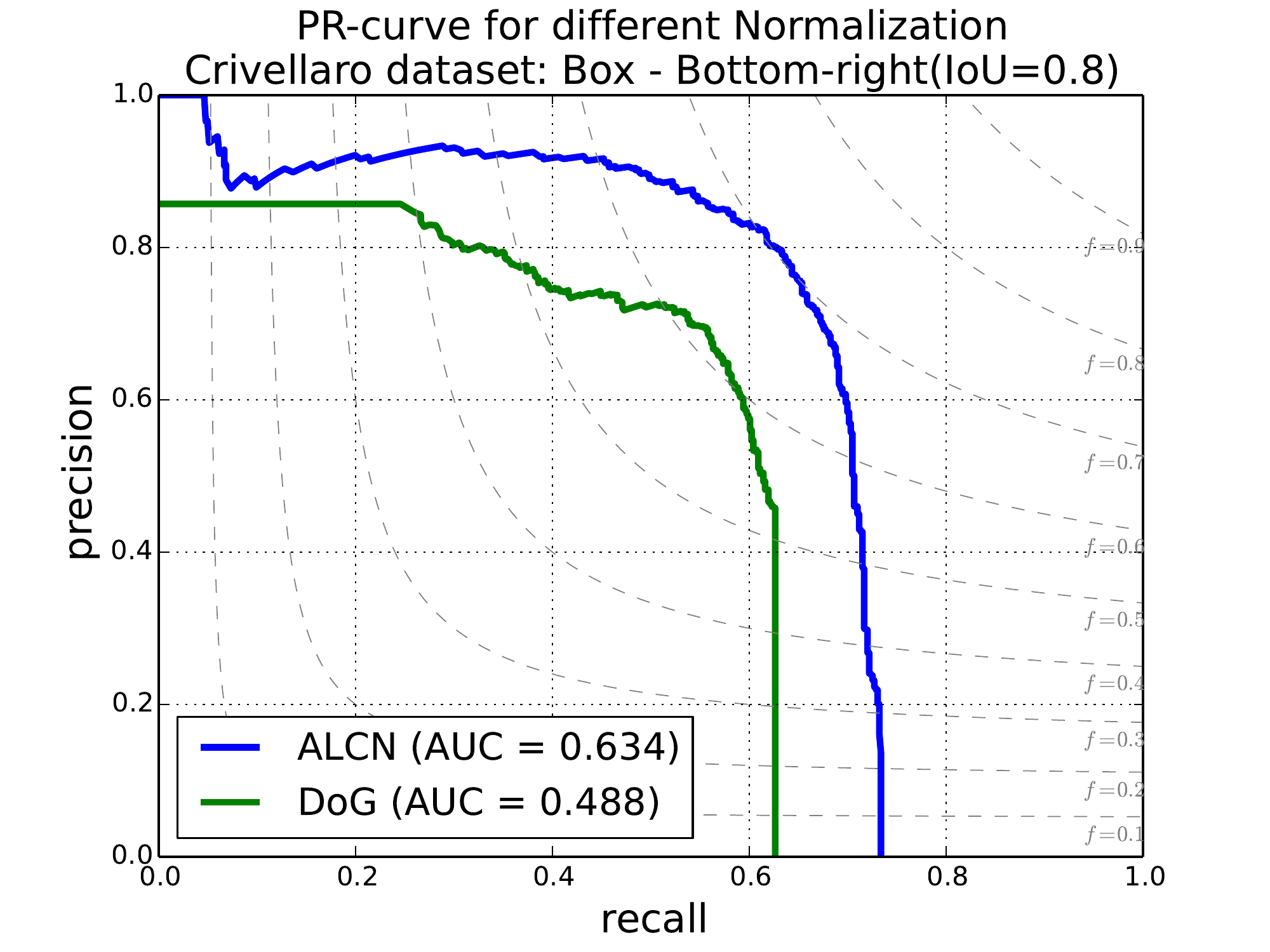} \\
    
    \end{tabular}
  \end{center}
  \vspace{-0.27cm}
  \caption{\label{fig:CrivellaroDetection}  Comparing ALCN  and DoG  on the  BOX
    dataset - Video \#3  from Crivellaro~(\cite{Crivellaro15}).  Our ALCN performs
    best at detecting the corners of the box.}
\end{figure}


\subsection{Image Normalization for Other Applications}

In this section, we evaluate our  normalization method on applications for which
it  was  not  trained for:  1)  3D  object detection, 2) 3D object pose  estimation, and 3) face  detection  and recognition.

\subsubsection{3D Object Pose Estimation}
\label{sec:3d_object_pose_estimation}

\begin{figure*}[t]
  \begin{center}
    \begin{tabular}{ccc}
      \includegraphics[width=0.3\linewidth]{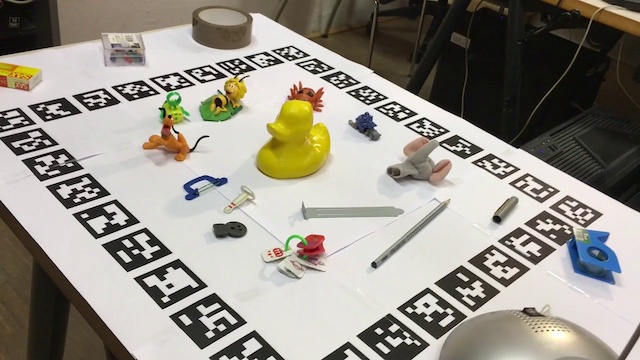} &
      \includegraphics[width=0.3\linewidth]{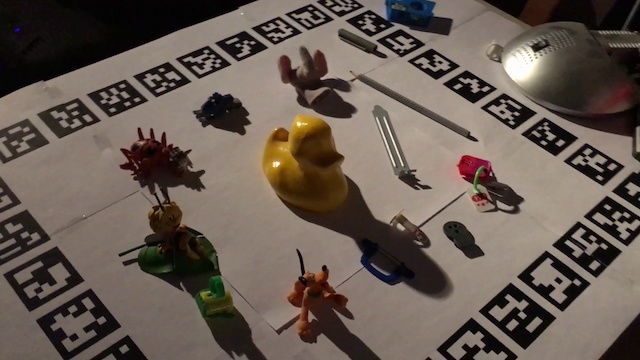} &
      \includegraphics[width=0.3\linewidth]{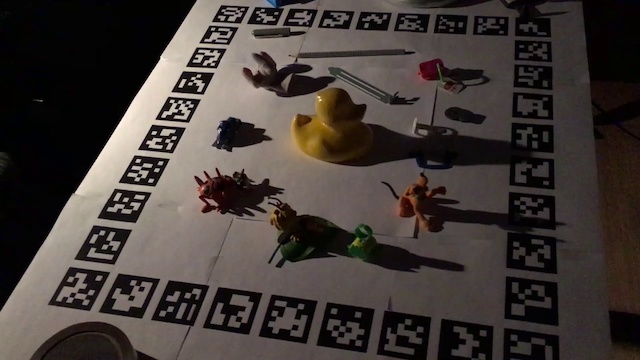}\\

      \includegraphics[width=0.3\linewidth]{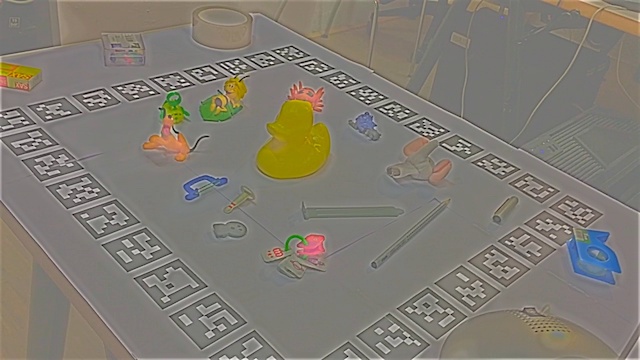} &
      \includegraphics[width=0.3\linewidth]{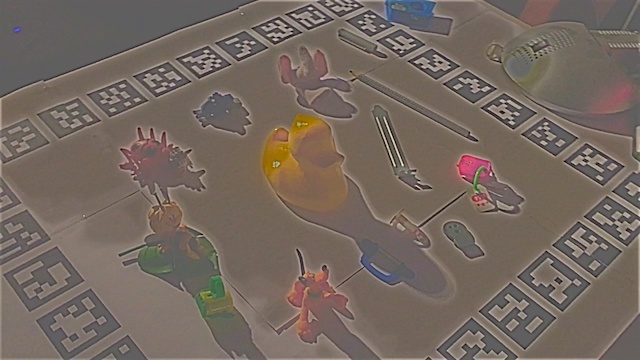} &
      \includegraphics[width=0.3\linewidth]{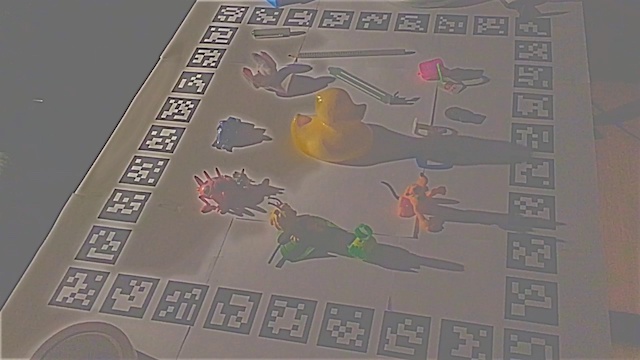}\\
      
    \end{tabular}
  \end{center}

  \caption{First row: Original  color images from the ALCN-Duck  dataset.    Second  row: Normalized color
    images after adding the ab channels of the  images on the first row to the normalized L
    channel of the images by our method.}
  \label{fig:imageNormalization_duck3d}
\end{figure*}

\newcommand{\ra}[1]{\renewcommand{\arraystretch}{#1}}

\begin{table*}\centering
  \ra{1.0}
  {\small 
\begin{tabular}{@{}lc|ccccccc@{}}
  \toprule
  & w/o {\footnotesize illumination changes} & \multicolumn{7}{c}{\footnotesize with illumination changes}\\
sequence     & \#1 & \#2 & \#3 & \#4 & \#5 & \#6 & \#7 & \#8 \\
\midrule
VGG      & {\bf 100} & 47.26      & 18.33       & 32.65    & 0.00  & 0.00  & 0.00 & 0.00\\
VGG+ALCN & {\bf 100} & {\bf 77.78} & {\bf 60.71} & {\bf 70.68} & {\bf 64.08 } & {\bf 51.37} & {\bf 76.20} & {\bf 50.10}\\
\bottomrule
\end{tabular}
}
\vspace{2mm}
\caption{\label{tbl:BB8_quantitative}  Percentage of  correctly estimated  poses
  using  the 2D  Projection metric  of  \cite{Brachmann16}, when  the method  of
  \cite{bb8}  is applied  to our  ALCN-Duck  sequences, with  and without  ALCN.
  Using  VGG---trained to  predict the  3D pose---alone  is not  sufficient when
  illumination changes. ALCN allows us to retrieve accurate poses.}
\end{table*}


As  mentioned in  the  introduction, our  main goal  is  to train  Deep
  Networks methods  for 3D pose  estimation, without requiring large   quantities of
  training data while being robust to  light changes.  We evaluate here ALCN for
  this goal on two different datasets.   

{\bf  BOX  dataset.}   To  evaluate  ALCN  for  3D  object  detection  and  pose
estimation,   we  first   applied   it   on  the   BOX   dataset  described   in
Section~\ref{sec:datasets}  using the  method of  \cite{Crivellaro15}, which  is
based on  part detection:  It first learns  to detect some  parts of  the target
object, then  it predicts the 3D  pose of each  part to finally combine  them to
estimate the object 3D pose.

The  test videos  from \cite{Crivellaro15}  exhibit challenging  dynamic complex
background  and light  changes. We changed the code provided  by the authors to
apply  ALCN before  the part  detection.   We evaluated  DoG normalization,  the
second best  method according  to our previous  experiments, optimized  on these
training  images,  against our  Normalizer.   Fig.~\ref{fig:CrivellaroDetection}
shows the results; ALCN allows us to  detect the parts more robustly and thus to
compute much more  stable poses.

{\bf ALCN-Duck  dataset.}  The method  proposed in~\cite{bb8} first  detects the
target object using a detector and then,  given the image window centered on the
object,  predicts the  3D  pose of  the  object using  a  regressor.  For  both,
detector and  regressor, (\cite{bb8}) finetunes convolution  and fully connected
layers of VGG~(\cite{Simonyan15}), and achieved very good results on the LineMOD
dataset.  However,  this dataset does not  exhibit strong light changes,  and we
evaluated    our   approach    on   the    ALCN-Duck   dataset    described   in
Section~\ref{sec:datasets}.  
Here, we use color  images as input to the detector
and the regressor.   To apply ALCN  to these images, we  use the method  proposed in
Section~\ref{sec:colorImageNormalization}. We normalized  color images by normalizing   the   L  channel   in   CIE   color   space. As our  experiments show  even if ALCN was trained on grayscale  images,  we get reasonably good normalized  color images. Fig.~\ref{fig:imageNormalization_duck3d} shows the normalized images  of the ALCN-Duck dataset.

Table~\ref{tbl:BB8_quantitative}  gives the  percentage  of correctly  estimated
poses using  the 2D  Projection metric~(\cite{Brachmann16})  with and  without our
ALCN  normalization.  \cite{bb8},  with and  without ALCN,  performs very  well on
video sequence \#1,  which has no illumination changes.  It  performs much worse
when ALCN is not used on Sequences \#2, \#3 and \#4, where the illuminations are
slightly different from training. For the  other sequences, which have much more
challenging lightening conditions, it dramatically fails to recover the object poses.
This shows that ALCN can provide illumination invariance at  a level to
which deep networks such as VGG cannot.  Some qualitative results are shown
on the last row of Fig.~\ref{fig:teaser}.


\subsubsection{Application to the Viola-Jones Detector}
In this  experiment, we  evaluate the performance  of the  Viola-Jones detection
algorithm  trained  with  a  training  set created  from  10  real  images,  and
normalized using  different methods.   

\begin{figure}
  \begin{center}
    \includegraphics[width=0.92\linewidth]{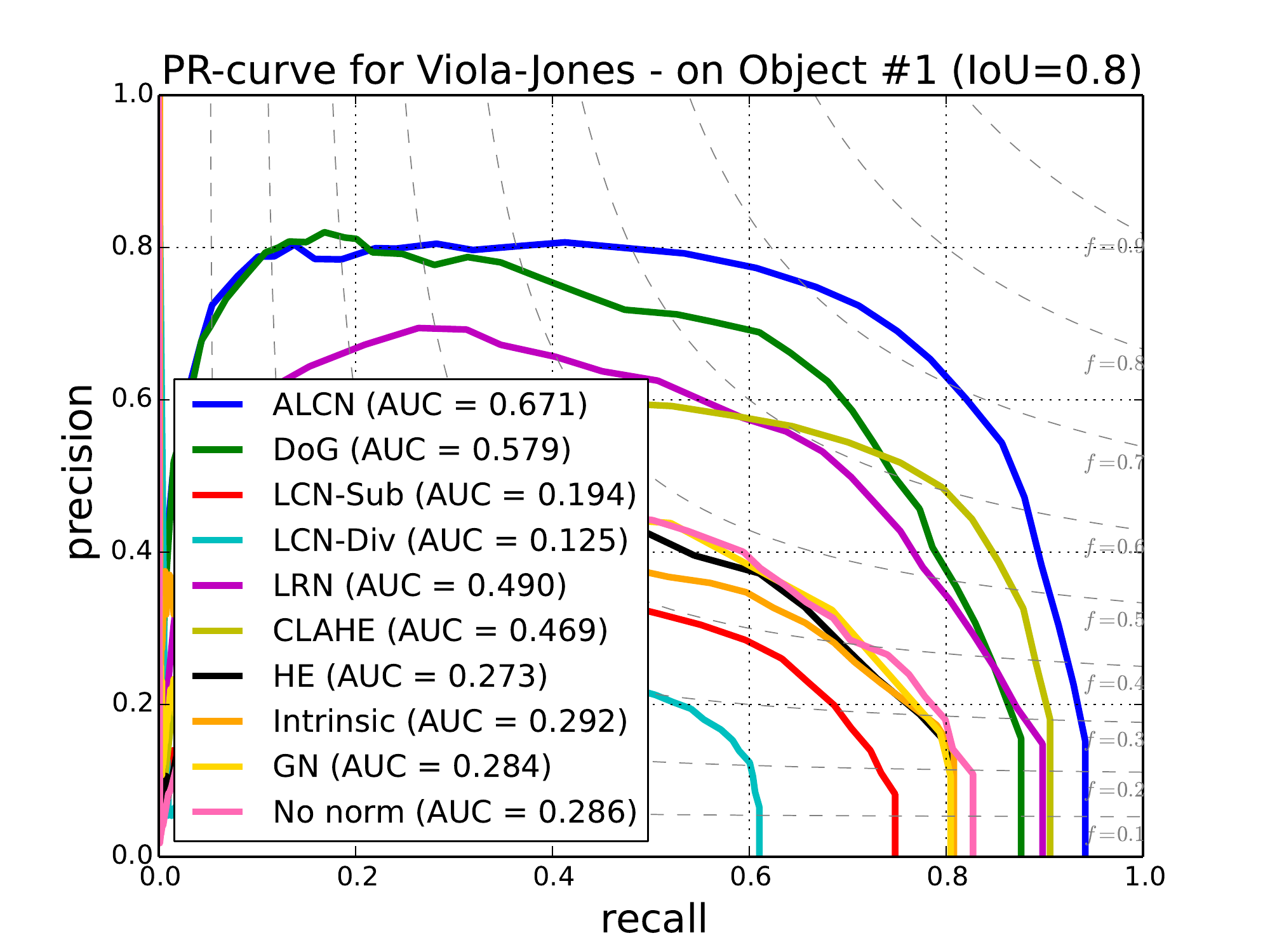}
  \end{center}
  \caption{\label{fig:ViolaJones}   Evaluating  different   normalization
      methods for the Viola-Jones detection method. By simply pre-processing the
      training and test images using our  ALCN, we can improve the performance of
      Viola-Jones detection with an  AUC from 0.286 to 0.671.  ALCN outperforms all the other
      normalization methods.}
\end{figure}

Fig.~\ref{fig:ViolaJones} shows that  Viola-Jones~(\cite{Viola04}) performs very
poorly with an AUC  of 0.286 in best cases.  However,  by simply normalizing the
training  and  test  images  using   our  ALCN,  Viola-Jones  suddenly  performs
significantly  better with  an  AUC  of 0.671,  while it  still does not  perform very well with other  normalization methods.   It may be  surprising that
Viola-Jones needs image normalization at all, as the Haar cascade image features
it relies on are very robust to light changes.  However, robustness comes at the
price of low discriminative power. With image normalization, the features do not
have to be as robust as they must be without it.

\subsubsection{Application to Face Recognition}

Finally we evaluate our  normalization for face recognition to see  if our normalization
can improve the  performance of current recognition algorithms.   Hence, we test
our normalization  on YaleBExt~(\cite{Georghiades:2001:FMI:378040.378083}) using
Eigenfaces~(\cite{Turk:1991:ER:1326887.1326894})                             and
Fisherfaces~(\cite{Belhumeur97}),  where  both  perform  poorly  with  different
normalization methods.  \cite{Han20131691} studied 13 different normalizations on
face recognition.   The best recognition  rates among the 13  normalizations are
59.3\% and 78.0\% VS 70.5\% and  98.6\% using our normalization, with Eigenfaces
and Fisherfaces respectively.

\section{Conclusion}

We proposed  an efficient approach  to illumination normalization, which improves
robustness to light changes for object  detection and 3D pose estimation methods
without requiring many training images.  We  have shown that our proposed method
can bring the power of Deep Learning  to applications for which large  quantities of
training  data  are  not  available,  since  it can  be  plugged  easily  to  other
applications.

\section*{Acknowledgments}
This work was supported by the Christian Doppler Laboratory for Semantic 3D Computer Vision, funded in part by Qualcomm Inc.

\bibliographystyle{model2-names}
\bibliography{string,vision}

\end{document}